%% file: main.tex
\documentclass{article}
\usepackage[utf8]{inputenc}
\usepackage{style}
\usepackage{tcolorbox}

\newcommand{\Oh}{\mathcal{O}}
\def\shownotes{1}  \ifnum\shownotes=1
\usepackage{todonotes}
\else
\usepackage[disable]{todonotes}
\fi
\newcommand{\pnote}[1]{\textcolor{magenta}{[PN: #1]}}
\newcommand{\jnote}[1]{\textcolor{cyan}{[JB: #1]}}
\newcommand{\dt}{\,\mathrm{d}t}
\newcommand{\supp}{\mathrm{supp}}

\newcommand{\smECE}{\textsf{smECE}}
\newcommand{\smECEc}{\smECE}
\newcommand{\smECEp}{\widetilde{\smECE}}
\newcommand{\ECE}{\textsf{ECE}}
\newcommand{\ce}{\mathrm{CE}}
\newcommand{\etace}{\ce_{\eta}}
\newcommand{\smce}{\mathrm{wCE}}
\newcommand{\ldce}{\underline{\mathrm{dCE}}}
\newcommand{\logit}{\mathrm{logit}}

\newcommand{\Bernoulli}{\mathrm{Bernoulli}}

\usepackage{algorithm2e}

\title{Smooth ECE:
Principled Reliability Diagrams\\via Kernel Smoothing}
\date{}

\author{
{\bf Jaros\l aw B\l asiok}\\
Columbia University
\and
{\bf Preetum Nakkiran}\\
Apple 
}

\usepackage{lineno}

\begin{document}

\maketitle

\begin{abstract}
\input{abstract}
\end{abstract}

\input{intro3}
\input{related}

\input{smoothECE}
\input{general-duality}
\input{experiments}
\input{conclusion}

\clearpage
\newpage

\bibliographystyle{plainnat}
\bibliography{calib_refs}
\appendix
\input{math}

\end{document}

%% file: abstract.tex
Calibration measures
and reliability diagrams
are two fundamental tools for measuring and interpreting
the calibration of probabilistic predictors.
Calibration measures quantify
the degree of miscalibration,
and reliability diagrams visualize the
structure of this miscalibration.
However, the most common constructions of reliability diagrams
and calibration measures --- binning and ECE ---
both suffer from well-known flaws
(e.g. discontinuity).
We show that a simple modification fixes both constructions:
first smooth the observations using an RBF kernel,
then compute the Expected Calibration Error (ECE) of this smoothed function.
We prove that with a careful choice of bandwidth,
this method yields a calibration measure that is
well-behaved in the sense of \citet*{UTC1}
--- a \emph{consistent calibration measure}.
We call this measure the \emph{SmoothECE}.
Moreover, the reliability diagram obtained 
from this smoothed function visually encodes
the SmoothECE, just as binned reliability diagrams encode
the BinnedECE.

We also provide a Python package with simple, hyperparameter-free
methods for measuring and plotting calibration:
\texttt{\`{}pip install relplot\`{}}.
Code at: \url{https://github.com/apple/ml-calibration}.


%% file: intro3.tex
\vspace{1cm}

\begin{figure}[h]
    \centering
    \includegraphics[width=0.9\textwidth]{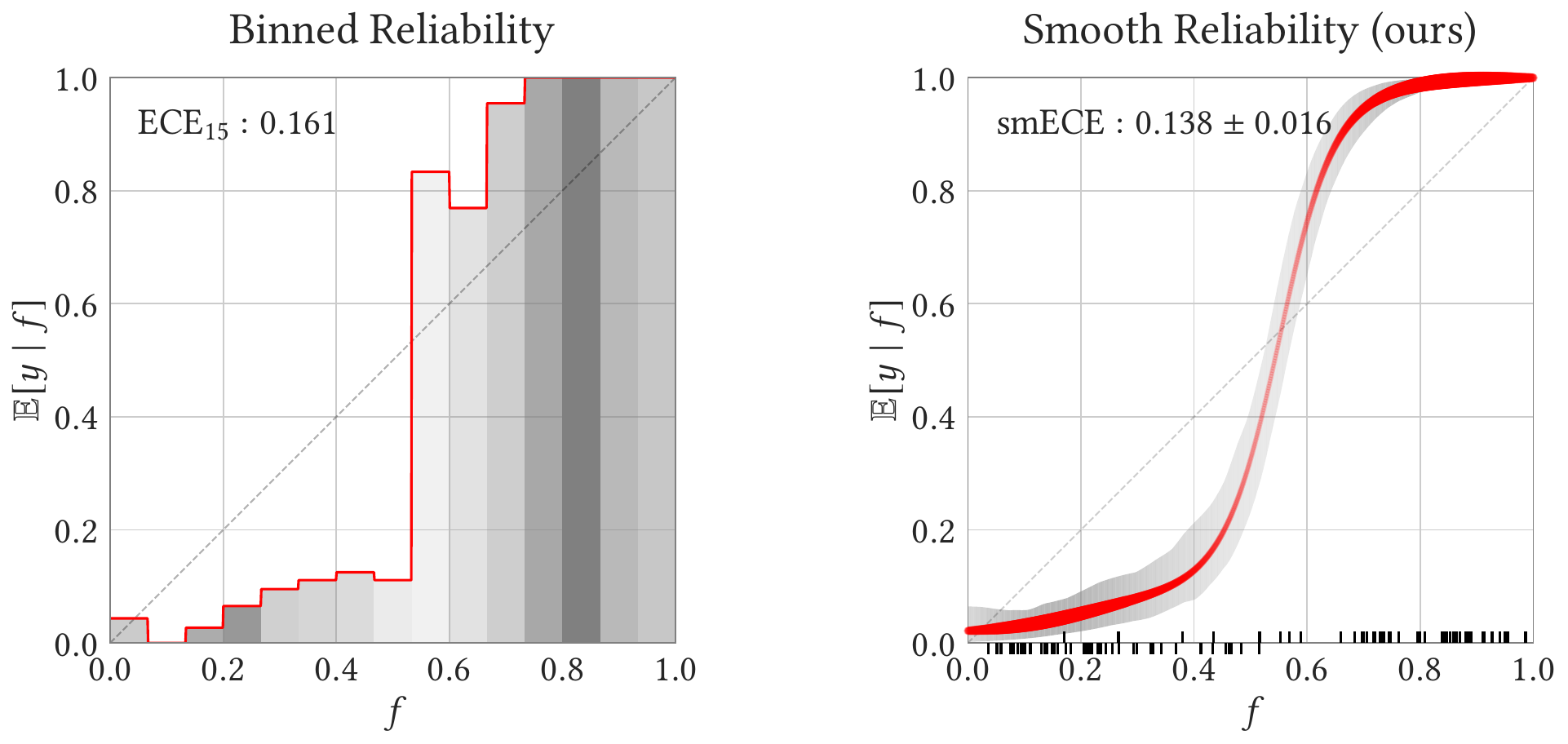}
    \caption{{\bf Left:} Traditional reliability diagram based on binning, which 
     is equivalent to histogram regression.
     {\bf Right:} Proposed reliability diagram based on kernel regression,
     with our theoretically-justified choice of bandwidth.
     The width of the red line corresponds to the density of predictions,
     and the shaded region shows bootstrapped confidence intervals.
     Plot generated by our Python package \texttt{relplot}.
     }
     \label{fig:hero}
\end{figure}

\newpage
\section{Introduction}
Calibration is a fundamental aspect of probabilistic predictors,
capturing how well predicted probabilities of events 
match their true frequencies \citep{dawid1982well}.
For example, a weather forecasting model is perfectly calibrated
(also called ``perfectly reliable'') if
among the days it predicts a 10\% chance of rain, the
observed frequency of rain is exactly 10\%.
There are two key
questions
in studying calibration:
First, for a given predictive model, how do we measure its
overall amount of miscalibration?
This is useful for ranking different models by their reliability,
and determining how much to trust a given model's predictions.
Methods for quantifying miscalibration are known as \emph{calibration measures}.
Second, how do we convey \emph{where} the miscalibration occurs?
This is useful for better understanding an individual predictor's behavior
(where it is likely to be over- vs. under-confident),
as well as for re-calibration--- modifying the predictor to make it better calibrated.
The standard way to convey this information is known as a \emph{reliability diagram}.
Unfortunately, in machine learning, the most common methods of constructing both calibration measures
and reliability diagrams suffer from well-known flaws,
which we describe below.

The most common choice of calibration measure in machine learning is the
Expected Calibration Error (ECE), more specifically its empirical variant the Binned ECE \citep{naeini2015obtaining}.
The ECE is known to be unsatisfactory for many reasons;
for example,
it is a discontinuous functional, so changing the predictor by an infinitesimally small
amount may change its ECE drastically \citep{kakadeF08,FosterH18,UTC1}.
Moreover, the ECE is impossible to estimate efficiently from samples \citep{lee2022t,arrieta2022metrics},
and its sample-efficient variant, the Binned ECE, is overly sensitive to choice of bin widths \citep{nixon2019measuring,kumar2019verified,minderer2021revisiting}.
These shortcomings have been well-documented in the community,
which motivated proposals of new, better-behaved calibration measures
(e.g. \citet{roelofs2022mitigating,arrieta2022metrics,lee2022t}).

Recently, \citet{UTC1} proposed a theoretical definition of
what constitutes a ``good'' calibration measure.
The key principle
is that good measures should provide upper and lower bounds
on the calibration distance $\ldce$,
which is the Wasserstein distance
between the joint distribution of prediction-outcome pairs,
and the set of perfectly calibrated such distributions
(formally defined in Definition~\ref{def:ldce} below).
Calibration measures
which satisfy this property
are called \emph{consistent calibration measures}.
In light of this line of work, one may think that
the question of which calibration measure to choose is largely resolved:
simply pick a consistent calibration measure,
such as Laplace Kernel Calibration Error / MMCE \citep{UTC1,kumar2018trainable},
as suggested by \citet{UTC1}.
However, this theoretical suggestion belies the practical reality:
Binned ECE remains the most popular calibration measure used in practice, even in recent studies.
We believe this is partly because Binned ECE enjoys an additional property:
it can be visually represented by a specific kind of reliability diagram, namely the binned histogram.
This raises the question of whether there are calibration measures
which are \emph{consistent} in the sense of \citet{UTC1},
and can also be represented by an appropriate reliability diagram.
To be precise, we must discuss reliability diagrams more formally.

\paragraph{Reliability Diagrams.}
We consider measuring calibration in the setting of binary outcomes, for simplicity.
Here, we have a joint distribution $(f, y) \sim \cD$
over predictions $f \in [0, 1]$, and true outcomes $y \in \{0, 1\}$.
We interpret $f$ as the predicted probability that $y=1$.
The ``calibration function''\footnote{In the terminology of \citet{brocker2008some}.} $\mu: [0, 1] \to [0, 1]$
is defined as the conditional expectation:
\[
\mu(f) := \E_{\cD}[ y \mid f ].
\]
A perfectly calibrated distribution, by definition,
is one with a diagonal calibration function: $\mu(f) = f$.
Reliability diagrams are traditionally thought of as 
estimates of the calibration function $\mu$ \citep{naeini2014binary,brocker2008some}.
In other words, \emph{reliability diagrams are one-dimensional regression methods}, since the goal of
regressing $y$ on $f$ is exactly to estimate the regression function $\E[y \mid f]$.
The practice of ``binning'' to construct reliability diagrams (as in Figure~\ref{fig:hero} left)
can be equivalently thought of as using histogram regression to regress $y$ on $f$.

With this perspective on reliability diagrams, one may wonder why
histogram regression is still the most popular method,
when more sophisticated regressors are available.
One potential answer is that users of reliability diagrams
have an additional desiderata:
it should be easy to visually 
read off a reasonable calibration measure
from the reliability diagram.
For example, it is easy to visually read off the Binned ECE
from a binned reliability diagram,
because it is simply the integrated absolute deviation from the diagonal:
\[
\mathrm{BinnedECE}_k
=
\int_0^1 \left| \hat{\mu}_k(f) - \bar{f}_k  \right| dF
\]
where $k$ is the number of bins, $\hat{\mu}_k$ is the histogram
regression estimate of $y$ given $f$, and
$\bar{f}_k$ is the ``binned'' version of $f$ --- 
formally the histogram regression estimate of $f$ given $f$.
This relationship is even more transparent for the full (non-binned) ECE,
where we have
\[
\mathrm{ECE}
=
\int_0^1 \left| \mu(f) - f  \right| dF
= \E_f[ |\mu(f) - f|]
\]
where $\mu$ is the true regression function as above.
However, more sophisticated regression methods do not neccesarily have such tight relationships to calibration measures.
Thus we have a situation where better calibration measures exist, but they are not accompanied by
reliability diagrams, and conversely better reliability diagrams exist (i.e. regression methods),
but they are not associated with consistent calibration measures.
We address this situation here:
we present a new
consistent calibration measure, \emph{SmoothECE}, 
along with a regression method which 
naturally encodes this calibration measure. 
The SmoothECE is, per its name, equivalent to the ECE of a
``smoothed'' version of the original distribution,
and the resulting reliability diagram
can thus be interpreted as a smoothed 
estimate of the calibration function.

We emphasize that the idea of smoothing is not new --- Gaussian kernel smoothing has been explicitly proposed
as method for constructing reliability diagrams in the past (e.g. \citet{brocker2008some},
as discussed in \citet{arrieta2022metrics}).
Our contribution is two-fold: first, we give strong theoretical justification 
for kernel smoothing by proving it induces a consistent calibration measure.
Second, and of more practical relevance, we 
show how to chose the kernel bandwidth in a principled way,
which differs significantly from existing recommendations.
In particular, in the past smoothing was recommended for
statistical reasons, to allow estimation of the calibration function
from finite samples. 
However, our analysis reveals that smoothing is necessary
for more fundamental reasons--- even if we have effectively infinite samples,
and a perfect estimate of the calibration function, we will still want to
use a non-zero smoothing bandwidth\footnote{Briefly, this is because
the true calibration function
still depends on the distribution $\cD$
in a discontinuous way. 
This discontinuity can manifest in the calibration measure,
unless it is appropriately smoothed.
}.
In other words, the smoothing is not done to approximate
some underlying population quantity-- rather, smoothing
is essential to the definition of the measure itself.




\subsection{Overview of Method \label{sec:method}}
We start by describing the regression method,
which defines our reliability diagram.
We are given i.i.d. observations $\{(f_1, y_1), (f_2, y_2) \ldots (f_k, y_k)\}$
where $f_i \in [0,1]$ is the $i$-th prediction, and $y_i \in \{0,1\}$ is the corresponding outcome.
For example, if we are measuring calibration of an ML model on a dataset
of validation samples,
we will have $f_i = F(x_i)$ for model $F$ evaluated on sample $x_i$, with ground-truth label $y_i$.
We would like to estimate the true calibration function 
$\mu(f) := \E[ f \mid y ]$.
Our estimate $\hat{\mu}(f)$ is given by Nadaraya-Watson kernel regression
(kernel smoothing) on this dataset (see \cite{nadaraya1964regression,watson1964regression,simonoff1996smoothing}):
\begin{equation}
\label{eqn:ksmooth}
    \hat{\mu}(f) := \frac{\sum_i {K}_{\sigma}(f, f_i) y_i}{\sum_i {K}_{\sigma}(f, f_i)}.
\end{equation}
That is, for a given $f\in [0,1]$ our estimate of $y$ is the weighted average of all $y_i$, where weights are given by the kernel function $K_\sigma(f, f_i)$.
The choice of kernel, and in particular the choice of bandwidth $\sigma$, is crucial
for our method's theoretical guarantees. We use an essentially standard kernel (described in more detail below): the Gaussian Kernel, reflected appropriately to handle boundary-effects of the interval $[0, 1]$.
Our choice of bandwidth $\sigma$ is more subtle, but it is not a hyperparameter -- 
we describe the explicit algorithm for choosing $\sigma$ in~\Cref{sec:smooth-ece}.
It suffices to say for now that the amount of smoothing $\sigma$ will end up being proportional
to the reported calibration error.

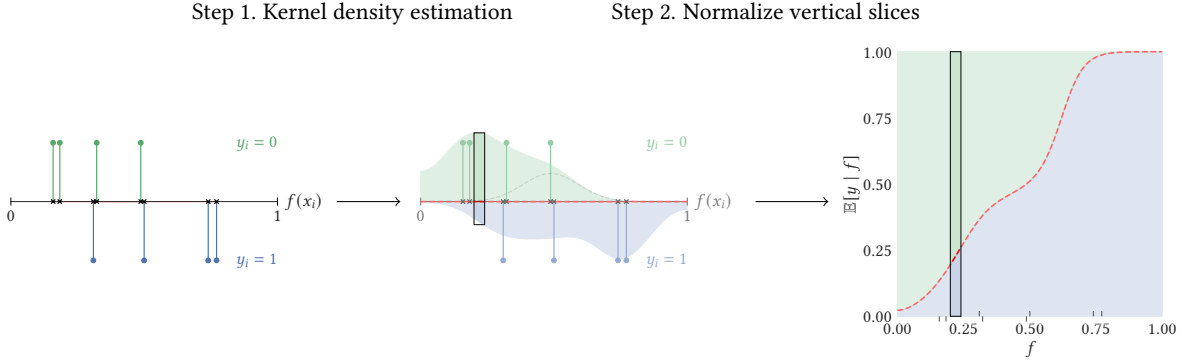
\begin{figure}[t]
    \centering
    \begin{tikzpicture}[scale=1.0]
\node (step0) at (0,0) {\includegraphics[scale=0.3]{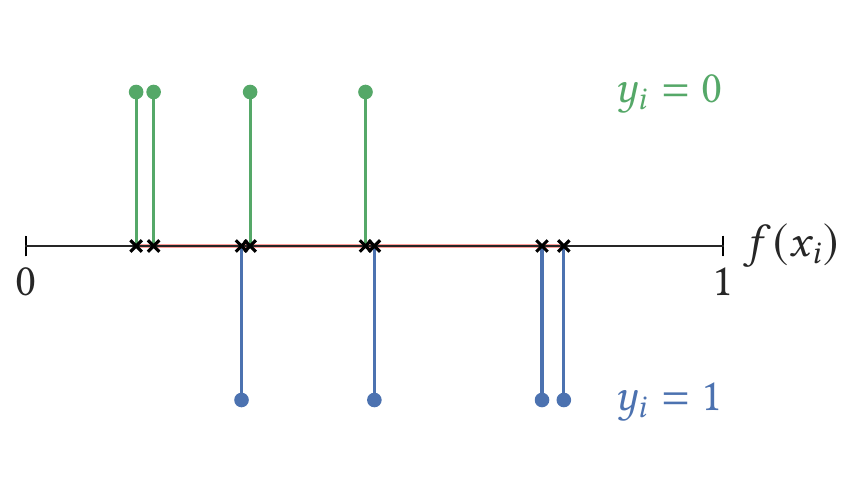}};
\node (step1_focus) at (5.5,0) {\includegraphics[scale=0.3]{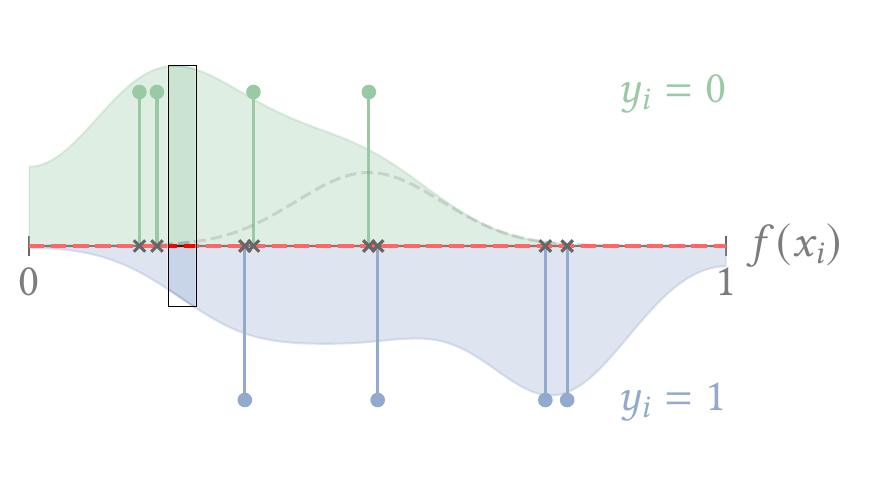}};
\node (step2) at (11.5,0) {\includegraphics[scale=0.3]{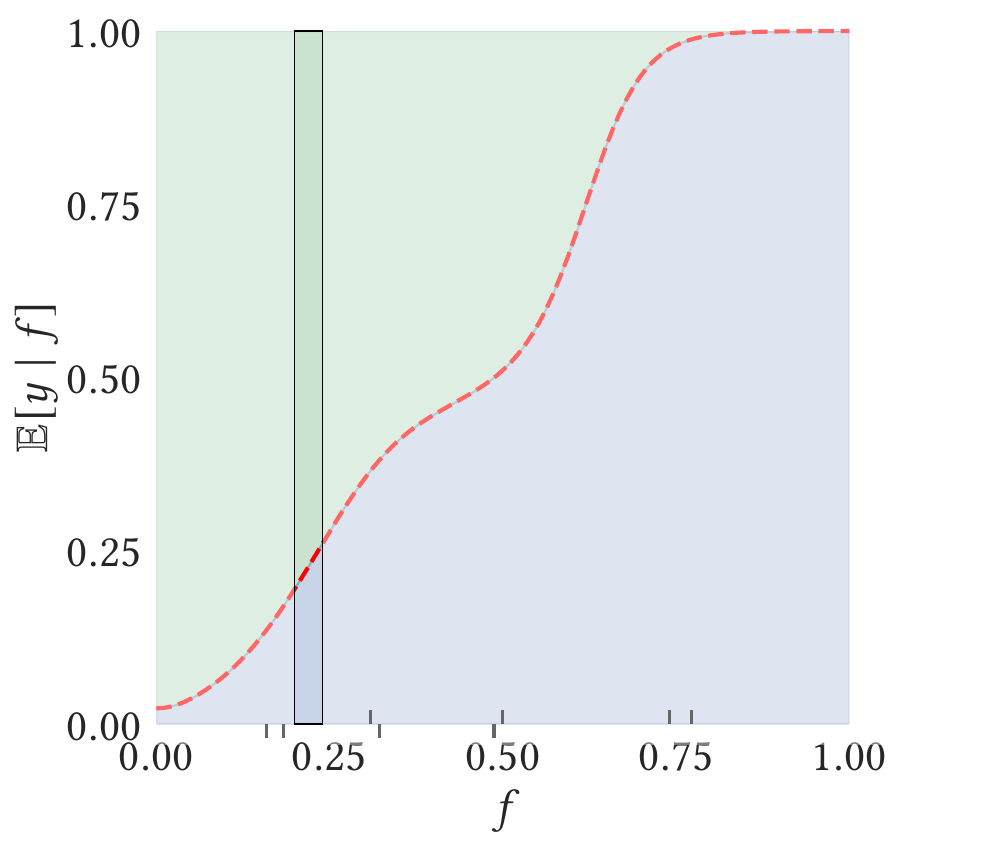}};
\draw[->] (step0) -- (step1_focus);
\draw[->] (step1_focus) -- (step2);
\node at (2.5, 2.5) {\small{Step 1. Kernel density estimation}};
\node at (8, 2.5) {\small{Step 2. Normalize vertical slices}};
\end{tikzpicture}
    \caption{
    \label{fig:kde}
    Illustration of how to compute the smooth reliability diagram,
    on a toy dataset of 8 samples.
    }
\end{figure}

An equivalent way of understanding the kernel smoothing of Eqn~\eqref{eqn:ksmooth}
is via kernel density estimation.
Specifically, let $\hat{\delta}_0$ and $\hat{\delta}_1$ be kernel density estimates of $p(f \mid y=0)$ and $p(f \mid y=1)$,
obtained by convolving the kernel $K_\sigma$ with the
empirical distributions of $f_i$ restricted to the samples where $y_i=0$ and $y_i=1$ accordingly.
Then it is easy to verify that $\hat{\mu}(f) = \hat{\delta}_1(f)/(\hat{\delta}_1(f) + \delta_2(f))$.
\Cref{fig:kde} illustrates this method of constructing the calibration function estimate $\hat{\mu}$,
on a toy dataset of eight prediction-outcome pairs $(f_i, y_i)$.

\paragraph{Reflected Gaussian Kernel}
In all of our kernel applications, we use a ``reflected'' version of the Gaussian kernel defined as follows.
Let $\pi_R :\mathbb{R} \to [0,1]$ be the projection function which is identity on $[0, 1]$, and collapses two points iff they differ by a composition of reflections around integers. That is $\pi_R(x) := (x \mod 2)$ if $(x \mod 2) \leq 1$, and $(2 - (x \mod 2))$ otherwise.
The Reflected Gaussian kernel on $[0, 1]$ with scale $\sigma$, is then given by 
\begin{equation}
\label{eq:reflected-gaussian}
    \tilde{K}_{\sigma}(x,y) := \sum_{\tilde{x} \in \pi_R^{-1}(x)} \phi_{\sigma}(\tilde{x} - y) = \sum_{\tilde{y} \in \pi_R^{-1}(y)} \phi_{\sigma}(x - \tilde{y}),
\end{equation}
where $\phi$ is the probability density function of $\mathcal{N}(0,1)$, that is $\phi_{\sigma}(t) = \exp(-t^2/2\sigma^2) / \sqrt{2 \pi \sigma^2}$.
Now, by construction, convolving a random variable $F$ with the kernel $\tilde{K}_{\sigma}$ 
produces the random variable $\pi_R(F + \eta)$ where $\eta\sim\mathcal{N}(0, \sigma)$.

We chose the Reflected Gaussian kernel
in order to alleviate the bias introduced by standard Gaussian kernel near the boundaries of the interval $[0, 1]$.
For instance, if we start with the uniform distribution over $[0,1]$,
convolve it with the standard Gaussian kernel,
and restrict the convolution to $[0,1]$,
we end up with a non-uniform distribution: the density close to the boundary is smaller by approximately factor of $2$ from the density in the middle.
In contrast, the reflected Gaussian kernel does not exhibit this pathology --- the uniform distribution on $[0,1]$ is an invariant measure under convolution with this kernel.

\paragraph{Reliability Diagram}
We then construct a reliability diagram in the standard way, by displaying
a plot of the estimated calibration function $\hat{\mu}$
along with a kernel density estimate of the predictions $f_i$ (see Figure~\ref{fig:hero}).
These two estimates, compactly presented on the same diagram,
provide a tool to quickly understand and visually assess calibration properties of a given predictor.
Moreover, they can be used to define a quantiative measure of overall degree of miscalibration, as we show below.


\paragraph{SmoothECE}
A natural measure of calibration can be easily computed from data in the above reliability diagram.
Specifically, let $\hat{\delta}(f)$ be the kernel density estimate of predictions:
$\hat{\delta}(f) := \frac{1}{n} \sum_i \tilde{K}_{\sigma}(f, f_i).$
Then, similar to the definition of ECE, we can integrate the deviation of $\hat{\mu}$ from the diagonal to obtain:

\begin{equation*}
    \smECEp_{\sigma} := \int |\hat{\mu}(t) - t| \hat{\delta}(t) dt.
\end{equation*}

The measure of calibration we actually propose, $\smECE_\sigma$, is closely related but not identical to the above.
Briefly, to define $\smECE_\sigma$ we consider the kernel smoothing of the difference between the outcome and the prediction $(y_i - f_i)$ instead of just smoothing the outcomes $y_i$. As it turns out, those choices lead to a calibration measure with better mathematical properties: $\smECE_{\sigma}$ is monotone decreasing as the kernel bandwidth $\sigma$ is increased,
and $\smECE$, applied to the population distribution, is 0 for perfectly calibrated predictors.


We reiterate that the choice of the scale $\sigma$ is very important: too large or too small bandwidth will
prevent the SmoothECE from being a consistent calibration measure.
In~\Cref{sec:smooth-ece}, we will show how to algorithmically define the correct scale $\sigma^*$.
For the reliability diagram, we suggest presenting the estimates $\hat{y}$ and $\hat{\delta}$ with the same scale $\sigma^*$, and for this scale we indeed have $\smECEp_{\sigma^*} \approx \smECE_{\sigma^*}$
(see~\Cref{sec:reliability-diagram}).
Finally, note that we have been discussing finite-sample estimators of all quantities;
the corresponding population quantities are defined analogously in Section~\ref{sec:smooth-ece}.


\subsection{Extensions to General Metrics}
Until now, we have been discussing the
original notion of
consistent calibration measure as introduced in \citet{UTC1}.
This notion relies on the concept of the \emph{distance to calibration} $\ldce$,
which we vaguely referred to before.
It turns out that
the notion of calibration distance, as a Wasserstein distance (the definition of Wasserstein distance is provided for readers convenience in \Cref{sec:smooth-ece}),
implicitly assumes the trivial metric on the space of predictions: for $f_1, f_2 \in [0,1]$, we consider $\ell_1(f_1, f_2) := |f_1 - f_2|$. In fact the associated distance to calibration can be defined generally for any metric.
Let us finally provide a formal definition of distance to calibration here.
\begin{definition}[Distance to Calibration]
    \label{def:ldce-d}
For a probability distribution $\cD$ over $[0,1] \times \{0,1\}$, and a metric $d : [0,1]^2 \to \R_{\geq 0} \cup \{+\infty\}$, we define $\ldce_d(\cD)$ to be the Wasserstein distance to the nearest perfectly calibrated distribution, with respect to the metric\footnote{This definition differs slightly from the original definition in \cite{UTC1} in that the distance on the second coordinate between two different outcomes was infinite  (see \Cref{def:ldce}). \Cref{clm:definitions-equivalent} shows that the specialization of this new definition to the trivial metric is equivalent up to a constant factor with the original $\ldce$, but the definition presented here behaves better in general --- it allows to generalize the duality (\Cref{thm:duality}) to arbitrary metrics on predictions $[0,1]$. } \begin{equation*}
        d_{[0,1]\times \{0,1\}}((f_1, y_1), (f_2, y_2)) := d(f_1, f_2) + |y_1 - y_2|.
    \end{equation*}
\end{definition}
There is indeed a good reason to consider \emph{non-trivial}
metrics on the space of predictions, 
because such metrics arise naturally when
minimizing a generic proper loss function.
This connection is well-known part of the convex analysis theory 
\citep{gneiting2007probabilistic},
and \citet{post-processing} builds upon this in the context of calibration error.
We summarize the latter results below.

It was shown in \citet{post-processing} that for the standard $\ell_1$ metric,
the $\ldce_{\ell_1}$ provides a lower and upper bound on how much the $\ell_2$-loss of a predictor can be improved by post-composition with a Lipschitz function.
Specifically, they showed that a random pair $(f,y) \in [0,1] \times \{0,1\}$ of prediction $f$ and outcome $y$ has small $\ldce_{\ell_1}$, if and only if the loss $\E (f-y)^2$ cannot be improved by more than $\eps$
by Lipschitz post-processing of the prediction.
That is, for any Lipschitz function $w: [0,1] \to [0,1]$, we have
\begin{equation*}
    \E (w(f) - y)^2 \geq \E (f-y)^2 - \varepsilon.
\end{equation*}
In practice, though,
prediction algorithms are often optimized
for different proper loss functions than $\ell_2$--- the cross-entropy loss is a popular choice
in deep learning, for example, leading to a metric $d_{\textrm{logit}}(u,v) := |\log(u/(1-u)) - \log(v/(1-v))|$.
With this in mind, \citet{post-processing} generalize both the notion of calibration error, and their theorem, to apply to general metrics.
This motivates a study of calibration error for a non-trivial metrics on the space of predictions~$[0,1]$, which we will undertake in the remainder of this paper.

\subsection{Summary of Our Contributions}

\begin{enumerate}
    \item {\bf SmoothECE.} We define a new hyperparmeter-free calibration measure, which we call the \emph{SmoothECE}
    (abbreviated $\smECE$).
    This measure 
We prove that the SmoothECE is a \emph{consistent calibration measure}, in the sense of \citet{UTC1}.
It also corresponds to a natural notion of distance:
if SmoothECE is $\eps$, then the function $f$ can be stochastically post-processed
to make it perfectly calibrated, without perturbing $f$ by more than $\eps$ in~$L_1$.

\item {\bf Smoothed Reliability Diagrams.}
We show how to construct principled reliability diagrams
which visually encode the SmoothECE.
These diagrams can be thought of as ``smoothed'' versions of the usual
binned reliability diagrams, where we perform Nadaraya-Watson kernel smoothing
with the Gaussian kernel.

\item {\bf Code.} We provide an open-source Python package \texttt{relplot}
which computes the SmoothECE
and plots the associated smooth reliability diagram.
It is hyperparameter-free, efficient, and includes
uncertainty quantification via bootstrapping.
We include several experiments in Section~\ref{sec:experiments}, for demonstration purposes.


\item {\bf Extensions to general metrics.}
On the theoretical side, we investigate how far our construction of SmoothECE generalizes. We show that the notion of SmoothECE introduced in this paper can indeed be defined for a wider class of metrics on the space of predictions $[0,1]$, and we prove the appropriate generalization of our main theorem: that
the $\smECE$ for a given metric is a consistent calibration measure with respect to the same metric.
Finally, perhaps surprisingly, we show that under specific conditions on the metric
(which are satisfied, for instance, by the $d_{\textrm{logit}}$ metric), the associated $\smECE$ is in fact a consistent calibration measure \emph{with respect to $\ell_1$ metric}.

\end{enumerate}

\paragraph{Organization.}
We begin by discussing the closest related works (Section~\ref{sec:related}).
In Section~\ref{sec:smooth-ece} we formally define the SmoothECE and prove its
mathematical and computational properties.
We provide the explicit algorithm (Algorithm~\ref{alg:smECE}),
and prove it is sample-efficient (Section~\ref{sec:samp-eff})
and runtime efficient (Section~\ref{sec:runtime}).
We then discuss the justification behind our various design choices
in Section~\ref{sec:reliability-diagram}, primarily to aid intuition.
Section~\ref{sec:gen-metrics} explores extensions of our results
to more general metrics.
Finally, we include experimental demonstrations of our method
and the associated python package in Section~\ref{sec:experiments},
and conclude in Section~\ref{sec:conclusion}.

%% file: related.tex
\section{Related Works}
\label{sec:related}

{\bf Reliability Diagrams and Binning.}
Reliability diagrams, as far as we are aware,
had their origins in the early reliability tables constructed by the meteorological community. \citet{hallenbeck1920forecasting}, for example, presents the performance of a certain rain forecasting method
by aggregating results over 6 months into a table:
Among the days forecast to have between $10\%-20\%$ chance of rain,
the table records the true fraction of days which were rainy ---
and similarly for every forecast interval.
This early account of calibration already applies the practice of binning--- discretizing predictions
into bins, and estimating frequencies conditional on each bin.
Plots of these tables turned into binned reliability diagrams \citep{murphy1977reliability,degroot1983comparison},
which was recently popularized in the machine learning community by
a series of works including \citet{zadrozny2001obtaining,niculescu2005predicting,guo2017calibration}.
Binned reliability diagrams continue to be used in studies
of calibration in machine learning, including in the GPT-4 tech report
\citep{guo2017calibration,nixon2019measuring,minderer2021revisiting,desai2020calibration,clarte2023expectation,openai2023gpt4}.

{\bf Reliability Diagrams as Regression.}
The connection between reliability diagrams and regression methods
has been noted in the literature (e.g. \citet{brocker2008some,copas1983plotting, stephenson2008two}).
For example, \citet{stephenson2008two} observes
``one can consider binning to be a crude form of non-parametric smoothing.''
However, this connection does not appear to be 
appreciated in the machine learning community,
since many seemingly unresolved questions about reliability diagrams
are clarified by the connection to statistical regression.
For example, there is much debate about how to choose hyperparameters when constructing
reliability diagrams via binning (e.g. the bin widths, the adaptive vs. non-adaptive binning scheme, etc), and it is not apriori clear how to think about the effect of these choices.
The regression perspective offers insight here: optimal hyperparameters 
in regression are chosen to minimize test-loss (e.g. on some held-out validation set). 
And in general, the choice of estimation method for reliability diagrams
should be informed by our assumptions and priors
about the underlying ground-truth calibration function $\mu$, such as smoothness
and monotonicity, just as it is in statistical regression.

Finally, we remind the reader of a subtlety: our objective
in this work is \emph{not}
identical to the regression objective, since we want an estimator
that is simultaneously a reasonable regression and a consistent calibration measure.
Our choice of bandwidth must thus
carefully balance the two; it cannot be simply be chosen to minimize the regression test loss.

{\bf Alternate Constructions.}
There have been various proposals to construct reliability diagrams which improve on binning;
we mention several of them here.
Many proposals can be seen as suggesting alternate regression techniques,
to replace histogram regression.
For example, some works suggest modifications to improve the binning method, such as 
adaptive bin widths or debiasing \citep{kumar2019verified, nixon2019measuring,roelofs2022mitigating}.
These are closely related to data-dependent histogram estimators
in the statistics literature \citep{nobel1996histogram}.
Other works suggest using entirely different regression methods,
including spline fitting \citep{guptacalibration},
kernel smoothing \citep{brocker2008some,popordanoska2022consistent}, and
isotonic regression \citep{dimitriadis2021stable}.
The above methods for constructing regression-based reliability diagrams
are closely related to methods for re-calibration, since the ideal recalibration
function is exactly the calibration function $\mu$.
For example, isotonic regression \citep{barlow1972statistical} has been used as both
for recalibration \citep{zadrozny2002transforming,naeini2015obtaining}
and for reliability diagrams \citep{dimitriadis2021stable}.
Finally, \citet{tygert2020plots} and \citet{arrieta2022metrics}
suggest visualizing reliability via cumulative-distribution plots,
instead of estimating conditional expectations.
While all the above proposals do improve upon binning in certain aspects,
none of them ultimately induce consistent calibration measures in the sense of \citet{UTC1}.
For example, the kernel smoothing proposals often suggest picking the kernel bandwidth to optimize the regression test loss. This choice does not yield a consistent
calibration measure as the number of samples $n \to \infty$.
See \citet{UTC1} for further discussion on the shortcomings of these measures.

{\bf Multiclass Calibration.}
We focus on binary calibration in this work.
The multi-class setting introduces several new complications--- foremost,
there is no consensus on how best to define calibration measures in the multi-class setting;
this is an active area of research (e.g. \citet{vaicenavicius2019evaluating,kull2019beyond,widmann2020calibration}).
The strongest notion of perfect calibration in the multi-class setting,
known as \emph{multi-class calibrated} or \emph{canonically calibrated},
is intractable to verify in general---
it requires sample-size exponential in the number of classes.
Weaker notions of calibration exist, such as classwise-calibration, 
confidence-calibration \citep{kull2019beyond}, and low-degree calibration \citep{GopalanKSZ22},
but it is unclear how to best define calibration measures
in these settings --- that is, how to most naturally extend the theory of \citet{UTC1} to multi-class settings.
We refer the reader to \citet{kull2019beyond} for a review of several different definitions of
multi-class calibration.

However, our methods can apply to specific multi-class settings which reduce to
binary problems.
For example, multi-class confidence calibration is equivalent to the
standard calibration of a related binary problem
(involving the joint distribution of confidences $f \in [0, 1]$ and accuracies $y \in \{0, 1\}$).
Thus, we can apply our method to plot reliability diagrams 
for confidence calibration, by first
transforming our multi-class data into its binary-confidence form.
We show an example of this in the neural-networks experiments in Section~\ref{sec:experiments}.

{\bf Consistent Calibration Measures.}
We warn the reader that the terminology of
\emph{consistent calibration measure}
does not refer to the concept of statistical consistency.
Rather, it refers to the definition introduced in \citet{UTC1},
to capture calibration measures
which polynomially approximate the true (Wasserstein) distance to 
perfect calibration.


%% file: smoothECE.tex
\section{Smooth ECE \label{sec:smooth-ece}}
In this section we will define the calibration measure $\smECE$. As it turns out, it has slightly better mathematical properties than $\smECEp$ defined in \Cref{sec:method}, and those properties will allow us to chose the proper scale $\sigma$ in a more principled way --- moreover, we will be able to relate $\smECE$ with $\smECEp$.

Specifically, the measure $\smECE_{\sigma}$
enjoys the following convenient mathematical properties,
which we will prove in this section.

\begin{itemize}
    \item The $\smECE_{\sigma}(\cD)$ is monotone decreasing
    as we increase the smoothing parameter $\sigma$.
    \item If $\cD$ is perfectly calibrated distribution, then for any $\sigma$ we have $\smECE_{\sigma}(\cD) = 0$. Indeed, for any $\sigma$ we have $\smECE_{\sigma}(\cD) \leq \ECE(\cD)$.
    \item The $\smECE_\sigma$ is Lipschitz with respect to the Wasserstein distance on the space of distributions over $[0,1] \times \{0,1\}$: for any $\cD_1, \cD_2$ we have $|\smECE_{\sigma}(\cD_1) - \smECE_{\sigma}(\cD_2)| \leq (1 + \sigma^{-1}) W_1(\cD_1, \cD_2)$. This implies $\smECE_{\sigma}(\cD) \leq (1 + \sigma^{-1})\ldce(\cD)$.
    \item For any distribution $\cD$ and any $\sigma$, there is a (probabilistic) post-processing $\kappa$, such that if $(f,y) \sim \cD$, then the distribution $\cD'$ of $(\kappa(f), y)$ is perfectly calibrated, and moreover $\E | f - \kappa(f)| \leq \smECE_{\sigma}(\cD) + \sigma$. In particular $\ldce(\cD) \leq \smECE_{\sigma} + \sigma$.
    \item Let us denote by $\cD_{\sigma}$ the distribution of $(\pi_R(f+\sigma \eta), y)$ for $(f,y) \sim \cD$ and $\eta \sim \mathcal{N}(0,1)$. (See~\eqref{eq:reflected-gaussian} for a definition of the wrapping function $\pi_R$.) Then
    $|\smECE_{\sigma}(\cD) - \ECE(\cD_\sigma)| < 0.8 \sigma$. In particular, for $\sigma_*$ such that $\smECE_{\sigma_*}(\cD) = \sigma_*$ we have $\smECE_{\sigma_*}(\cD) \approx \ECE(\cD_{\sigma_*}).$
\end{itemize}

\subsection{Defining $\smECE_{\sigma}$ at scale $\sigma$}
We now present the construction
of $\smECE_{\sigma}$, at a \emph{given} scale $\sigma > 0$.
We will show how to pick this $\sigma$ in the subsequent section.
Let $\cD$ be a distribution over $[0,1] \times \{0, 1\}$ of the pair of prediction $f \in [0,1]$ and outcome $y \in \{0, 1\}$. For a given kernel $K:\R \to \mathbb{R}$ we define the kernel smoothing of the residual $r := y-f$ as
\begin{equation}
    \hat{r}_{\cD, K}(t) := \frac{\E_{(f,y) \sim \cD} K(t, f) (y-f)}{\E_{(f,y) \sim \cD} K(t,f)}.
    \label{eq:r_hat}
\end{equation}
This differs from the definition in \Cref{sec:method}, where we applied the kernel smoothing to the outcomes $y$ instead.

In many cases of interest, the probability distribution $\cD$ is going to be an empirical probability distribution over finite set of pairs $\{(f_i, y_i)\}$ of observed predictions $f_i$ and associated observed outcomes $y_i$. In this case, the $\hat{r}_{\cD}(t)$ is just a weighted average of residuals $(f_i - y_i)$ where the weight of a given sample is determined by the kernel $K(f_i, t)$. This is equivalent to the Nadaraya-Watson kernel regression (a.k.a. kernel smoothing, see \cite{nadaraya1964regression,watson1964regression,simonoff1996smoothing}), estimating $(y-f)$ with respect to the independent variable $f$. 

We consider also the kernel density estimation
\begin{equation}
    \hat{\delta}_{\cD, K}(t) := \E_{f,y \sim \cD} K(t, f). \label{eq:delta_hat}
\end{equation}
Note that if the kernel $K$ is of form $K(u,v) = K(u-v)$ where the univariate $K$ is a probability density function of a random variable $\eta$, then $\hat{\delta}$ is a probability density function of the random variable $\eta + f$ (with $\eta, f$ independent), and moreover we can interpret $\hat{r}_K(t)$ as $\E[y-f | f + \eta = t]$.

The $\smECEc_K(\cD)$ is now defined as
\begin{equation}
    \smECEc_K(\cD) := \int |\hat{r}_{\cD, K}(t)| \hat{\delta}_{\cD, K}(t) \dt.
    \label{eq:smooth-ece}
\end{equation}

This notion is close to $\ECE$ of a smoothed distribution of $(f,y)$. We will provide more rigorous guarantees in~\Cref{sec:reliability-diagram}. For now, let us discuss the intuitive connection. For any distribution of prediction, and outcome $(f, y)$, we can consider an expected residual $r(t) := \E [ f-y | f = t]$, then 
\begin{equation*}
    \ECE(f,y) := \int |r(t)| \, \mathrm{d} \mu_{f}(t),
\end{equation*}
where $\mu_f$ is a measure of $f$. We can compare this with~\eqref{eq:smooth-ece}, where the conditional residual $r$ has been replaced by its smoothed version $\hat{r}$, and the measure $\mu_f$ has been replaced by $\hat{\delta} \dt$ -- the measure of $f + \eta$ for some noise $\eta$.

The equation~\eqref{eq:smooth-ece} can be simplified by directly combining the equations~\eqref{eq:r_hat} and~\eqref{eq:delta_hat},
\begin{equation}
    \smECEc_K(\cD) = \int \left|\E_{f,y} K(t, f) (y-f)\right| \dt.
    \label{eq:smooth-ece-integral}
\end{equation}

In what follows, we will be focusing on the reflected Gaussian kernel with scale $\sigma$, $\tilde{K}_{N, \sigma}$ (see~\eqref{eq:reflected-gaussian}), and we shall use shorthand $\smECEc_\sigma$ to denote $\smECEc_{\tilde{K}_{N, \sigma}}$. 
We will now show how the scale $\sigma$ is chosen.

\subsection{Defining $\smECE$: Proper choice of scale}
First, we observe that $\smECE_\sigma$ satisfies a natural monotonicity property:
increasing the smoothing scale $\sigma$ decreases the $\smECE_\sigma$.
(Proof of this and subsequent lemmas can be found in \Cref{app:proofs}.) 
\begin{lemma}
\label{lem:monotninicity-intro}
    For a distribution $\cD$ over $[0,1] \times \{0,1\}$ and $\sigma_1 \leq \sigma_2$, we have
    \begin{equation*}
        \smECEc_{\sigma_1}(\cD) \geq \smECEc_{\sigma_2}(\cD).
    \end{equation*}
\end{lemma}

Several of our design choices were crucial to ensure this property:
the choice of the reflected Gaussian kernel, and 
the choice to smooth the residual $(y-f)$ as opposed to the outcome $y$.


Note that since $\smECEc_{\sigma}(\cD) \in [0,1]$, and for a given predictor $\cD$, the function $\sigma \mapsto \smECEc_\sigma(\cD)$ is a non-increasing function of $\sigma$, there is a unique $\sigma^*$ s.t. $\smECEc_{\sigma^*}(\cD) = \sigma^*$ (and we can find it efficiently using binary search).

\begin{definition}[SmoothECE]
    We define $\smECE(\cD)$
    to be the unique $\sigma^*$ satisfying $\smECE_{\sigma^*}(\cD) = \sigma^*$.
    We also write this quantity as $\smECE_{*}(\cD)$ for clarity.
\end{definition}

\subsection{$\smECE$ is a consistent calibration measure}

We will show that $\sigma_*$ defined in the previous subsection is a convenient scale on which the $\smECEc$ of $\cD$ should be evaluated. The formal requirement that $\smECE_{\sigma^*}$ meets is captured by the notion of \emph{consistent calibration measure}, introduced in \cite{UTC1}. We provide the definition below, but before we do, let us recall the definition of the \emph{Wasserstein metric}.

For a metric space $(\cX, d)$, let us consider $\Delta(\cX)$ to be the space of all probability distributions over $\cX$. We define the \emph{Wasserstein} metric on the space $\Delta(X)$ (sometimes called \emph{earth-movers distance}) \cite{peyre2019computational}.

\begin{definition}[Wasserstein distance]
For two distributions $\mu, \nu \in \Delta(\cX)$ we define the Wasserstein distance
\begin{equation*}
    W_1(\mu, \nu) := \inf_{\gamma \in \Gamma(\mu,\nu)} \E_{(x,y) \sim \gamma} d(x,y),
\end{equation*}
where $\Gamma(\mu, \nu)$ is the family of all couplings of distributions $\mu$ and $\nu$.
\end{definition}

\begin{definition}[Perfect calibration]
A probability distribution $\cD$ over $[0,1]\times \{0,1\}$ of prediction $f$ and outcome $y$ is \emph{perfectly calibrated} if $\E_{\cD}[y|f]=f$. We denote the family of all perfectly calibrated distributions by $\mathcal{P} \subset \Delta([0,1] \times \{0, 1\})$.
\end{definition}
\begin{definition}[Consistent calibration measure \citep{UTC1}]
\label{def:ldce}
    For a probability distribution $\cD$ over $[0,1] \times \{0,1\}$ we define the distance to calibration $\ldce(\cD)$ to be the Wasserstein distance to the nearest perfectly calibrated distribution, associated with metric $d$ on $[0,1] \times \{0,1\}$ which puts two disjoint intervals infinitely far from each other.
    \footnote{This definition, which appeared in the \citep{UTC1} differs slightly from the more general definition \Cref{def:ldce-d} introduced in this work, in that \Cref{def:ldce-d} puts the two intervals within distance $1$ from each other, as opposed to $\infty$ in \Cref{def:ldce}. As we show with~\Cref{clm:definitions-equivalent}, this choice does not make substantial difference for the standard metric on the interval, but the \Cref{def:ldce-d} better generalizes to other metrics.}
    
    Concretely 
    \begin{equation*}
        d( (f_1, y_1), (f_2, y_2)) := 
        \begin{cases}
            |f_1 - f_2| & \text{if } y_1 = y_2 \\
            \infty &\text{otherwise}
        \end{cases}.
    \end{equation*}
    and
    \begin{equation*}
        \ldce(\cD) = \inf_{\cD \in \mathcal{P}} W_1(\cD, \cD').
    \end{equation*}
        
    Finally, any function $\mu$ assigning to distributions over $[0, 1]\times \{0, 1\}$ a non-negative real calibration score, is called \emph{consistent calibration measure} if it is polynomially upper and lower bounded by $\ldce$, i.e. there are constants $c_1, c_2, \alpha_1, \alpha_2$, s.t. 
    \begin{equation*}
        c_1 \ldce(\cD)^{\alpha_1} \leq \mu(\cD) \leq c_2\ldce(\cD)^{\alpha_2}.
    \end{equation*}
\end{definition}

With this definition in hand, we prove the following.
\begin{theorem}
    \label{thm:consistent-intro}
    The measure $\smECEc(\cD)$ is a consistent calibration measure.
\end{theorem}

This theorem is a consequence of the following two inequalities. First of all, if we add the penalty proportional to the scale of noise $\sigma$, then $\smECE_{\sigma}$ upper bounds the distance to calibration.
\begin{lemma}
\label{lem:lb-intro}
    For any $\sigma$, we have
    \begin{equation*}
        \ldce(\cD) \lesssim \smECEc_\sigma(\cD) + \sigma.
    \end{equation*}
\end{lemma}

On the other hand, as soon as the scale of the noise is sufficiently large compared to the distance to calibration, the $\smECEc$ of a predictor is itself upper bounded as follows.
\begin{lemma}
\label{lem:ub-intro}
Let $(f,y)$ be any predictor. Then for any $\sigma$ we have 
\begin{equation*}
    \smECEc_{\sigma}(\cD) \leq \left(1 + \frac{1}{\sigma}\right) \ldce(\cD).
\end{equation*}
In particular if $\sigma > 2\sqrt{\ldce(\cD)}$, then
\begin{equation*}
    \smECEc_{\sigma}(\cD) \leq 2 \sqrt{\ldce(\cD)}.
\end{equation*}
\end{lemma}
This lemma, together with the fact that $\sigma \mapsto \smECE_{\sigma}$ is non-increasing, directly implies that the fixpoint satisfies $\sigma^* \leq 2 \sqrt{\ldce(\cD)}$. On the other hand, using \Cref{lem:lb-intro}, at this fixpoint we have $\ldce(\cD) \leq \sigma^* + \smECE_{\sigma^*}(\cD) = 2 \sigma^*$. That is
\begin{equation*}
    \frac{1}{2} \ldce(\cD) \leq \smECE_*(\cD) \leq 2 \sqrt{\ldce(\cD)},
\end{equation*}
proving the \Cref{thm:consistent-intro}.

\begin{remark}
    We wish to clarify the significance of the decision to use the \emph{reflected Gaussian kernel} as a kernel $K$ of choice in the definition of $\smECE$. 

    Upper and lower bounds \Cref{lem:lb-intro} and \Cref{lem:ub-intro} hold for a wide range of kernels, and indeed, we prove more general statements (\Cref{lem:lb} and \Cref{lem:ub}) in the appendix. In order to deduce the existance of unique $\sigma^*$ we need the monotonicity property (\Cref{lem:monotninicity-intro}), which is more subtle. For this property to hold, we require that for $\sigma_1 \leq \sigma_2$, we can decompose the kernel $K_{\sigma_2}$ as a convolution: $\tilde{K}_{\sigma_2} = \tilde{K}_{\sigma_1} \ast \tilde{K}_{h(\sigma_1, \sigma_2)}$. This property is also satisfied for instance for a standard Gaussian kernel on $K_{N, \sigma} : \R \times \R \to \mathbb{R}$, given by $K(x,y) = \phi_{\sigma}(x-y)$ --- and indeed, we could have stated (and proved) \Cref{thm:consistent-intro} for this kernel. In fact, in~\Cref{app:proofs} we showed all necessary lemmas in the generality that covers also this case. 

    The choice of reflected Gaussian kernel, instead of Gaussian kernel stems from the fact, that the domain of reflected Gaussian kernel is $[0,1]$ instead of $\R$ --- more natural choice, since our distribution is indeed supported in $[0,1]$. The standard Gaussian kernel would introduce undesirable biases in the density estimation near the boundary of the interval $[0,1]$ --- a region which is of particular interest. The associated reliability diagrams would therefore be less informative --- for instance, the uniform distribution on $[0,1]$ is invariant under convolution with our reflected Gaussian kernel, which is not the case for the standard Gaussian kernel. 
\end{remark}

\subsection{Sample Efficiency}
\label{sec:samp-eff}
We show that we can estimate $\smECE$ of the underlying distribution $\cD$ over $[0,1] \times \{0,1\}$ (of prediction $f \in [0,1]$ and outcome $y \in \{0,1\}$), using few samples from this distribution. Specifically, let us sample independently  at random $m$ pairs $(f_i, y_i) \sim \cD$, and let us define $\hat{\cD}$ to be the empirical distribution over the multiset $\{ (f_i, y_i) \}$; that is, to sample from $\hat{\cD}$, we pick a uniformly random $i \in [m]$ and output $(f_i, y_i)$. 
\begin{theorem}
\label{thm:sample-complexity}
For a given $\sigma_0 > 0$
if $m \gtrsim \sigma_0^{-1} \varepsilon^{-2}$, then with probability at least $2/3$ over the choice of independent random sample $(f_i, y_i)_{i=1}^m$ (with $(f_i, y_i) \sim \cD$), we have simultanously for all $\sigma \geq \sigma_0$,
\begin{equation*}
    |\smECE_{\sigma}(\cD) - \smECE_{\sigma}(\hat{\cD})| \leq \varepsilon.
\end{equation*}

In particular if $\smECE_*(\cD) > \sigma_0$, then (with probability at least $2/3$) we have $|\smECE_{*}(\cD) - \smECE_{*}(\hat{\cD})| \leq \varepsilon.$
\end{theorem}
The proof can be found in~\Cref{app:sample-complexity}.
The success probability can be amplified in the standard way,
by taking the median of independent trials.

\subsection{Runtime}
\label{sec:runtime}
In this section we discuss how $\smECE$ can be computed efficiently: for a given sample $\{(f_1, y_1) \ldots (f_n, y_n)\} \in ([0,1]\times\{0,1\})^n$, if $\hat{\cD}$ is an empirical distribution over $\{(f_i, y_i)\}_{i=1}^n$, then the quantity $\smECE_{\sigma}(\hat{\cD})$ can be approximated up to error $\varepsilon$ in time $\Oh(n + M^{-1} \log^{3/2} M^{-1})$ in the RAM model, where $M = \lceil \varepsilon^{-1} \sigma^{-1} \rceil$. In order to find an optimal scale $\sigma_*$, we need to perform a binary search, involving $\log \varepsilon^{-1}$ evaluations of $\smECE_{\sigma}$. We provide the pseudocode as Algorithm~\ref{alg:smece_sigma} for computation of $\smECE_{\sigma}$ on a given scale $\sigma$ and Algorithm~\ref{alg:smECE} for finding the $\sigma_*$.

We shall first observe that the convolution with the reflected Gaussian kernel can be expressed in terms of a convolution with a shift-invariant kernel. This is useful, since such a convolution can be implemented in time $\Oh(M \log M)$ using Fast Fourier Transform, where $m$ is the size of the discretization.
\begin{claim}
\label{clm:reflected-kernel-wrapping}
For any function $g : [0,1] \to \R$, the convolution with the reflected Gaussian kernel $g \ast K_{N, \sigma}$ can be equivalently computed as follows. Take an extension of $g$ to the entire real line $\tilde{g} : \R \to \R$ defined as $\tilde{g}(x) := g(\pi_R)(x)).$
Then
\begin{equation*}
    [g \ast \tilde{K}_{N, \sigma}](t) = [\tilde{g} \ast K_{N, \sigma}](t),
\end{equation*}
where $K_{N, \sigma} : \R \times \R \to \R$ is the standard Gaussian kernel $K_{\sigma}(t_1, t_2) = \exp(-(t_1 - t_2)^2/2\sigma)/\sqrt{2 \pi \sigma^2}$.
\end{claim}
\begin{proof}
Elementary calculation.
\end{proof}
We can now restrict $\tilde{g}$ to the interval $[-T, T+1]$ where $T := \lceil\sqrt{\log(2 \varepsilon^{-1})}\rceil$, convolve such a restricted $\tilde{g}$ with a Gaussian, and restrict the convolution in turn to the interval $\varepsilon$. Indeed, such a restriction introduces very small error, for every $t\in [0,1]$ we have.
\begin{align*}
     [(\mathbf{1}_{[-T, T+1]} \cdot \tilde{g}) \ast K_{N, \sigma}](t) - [\tilde{g} \ast K_{N, \sigma}](t) & \leq (1 - \Phi(T/\sigma)) + (1 - \Phi((T+1)/\sigma)) \\
    & \leq \sqrt{2/\pi}(T/\sigma) \exp(-(T/\sigma)^2/2).
\end{align*}

In practice, it is enough to reflect the function $g$ only twice, around two of the boundary points (corresponding to the choice $T=1$). For instance, when $\sigma < 0.38$, the above bound implies that the introduced additive error is smaller than $\sigma^2$, and the  error term rapidly improves as $\sigma$ is getting smaller. 

\RestyleAlgo{ruled}
\begin{algorithm}
\SetFuncSty{textrm}
\SetKwProg{Fn}{Function}{ is}{end}
\SetKwFunction{FDiscretization}{Discretization}
\Fn{\FDiscretization{$\{(f_i, z_i\}_{i=1}^n$, $M$}}{
$h \gets \textrm{zeros}(M+1)$\;
\For {$i \in [n]$}{
    $b \gets \textrm{round}(M f_i)$\;
    $h_b \gets h_b + z_i$\;
}
\textbf{return}\ $h$\;
}
\SetKwFunction{FWrap}{Wrap}
\Fn{\FWrap{h, T}}{
$M \gets \mathrm{len}(h)$\;
\For{$i \in [(2T+1)M]$}{
    $j \gets (i \mod 2 M)$\;
    \If{ $j > M$ }{
        $j \gets 2M - j$\; 
    }
    $\tilde{h}_i \gets h_j$\;
}
\textbf{return} $\tilde{h}$\;
}
\SetKwFunction{FDiscreteGaussianKernel}{DiscreteGaussianKernel}
\Fn{$\smECE(\sigma, \{f_i, y_i\}_1^n)$}{
$h \gets \FDiscretization(\{f_i, f_i - y_i\}, \lceil \sigma^{-1}\varepsilon^{-1}\rceil)$\;
$\tilde{h} \gets \FWrap(h, \lceil \sqrt{\log(2 \varepsilon^{-1})} )$\;
$K \gets \FDiscreteGaussianKernel(\sigma, \lceil \sigma^{-1}\varepsilon^{-1}\rceil)$\;
$\tilde{r} \gets \tilde{h} \ast K$\;
\textbf{return} $\sum_{i= T M}^{(T+1)M - 1} |\tilde{r}_i|$\;
}
\caption{Efficient estimation of $\smECE_\sigma$, at fixed scale $\sigma$} \label{alg:smece_sigma}
\end{algorithm}

\begin{algorithm}
\KwData{$(f_i, y_i)_{1}^n, \varepsilon$}
\KwResult{$\smECEc_*(\{(f_i, y_i)\})$}
$l \gets 0$\;
$u \gets 1$\;
\While {$u - l > \varepsilon$}{
    $\sigma \gets (u + l) / 2$\;
    \eIf{$\smECE_{\sigma}(\{f_i, y_i\}) < \sigma$}{
        $u \gets \sigma$\;
    }{
        $l \gets \sigma$\;
    }
}
$\textbf{return}\ u;$
\caption{Efficient estimation of $\smECE$: using binary search over $\sigma$ to find a root of the decreasing function $g(\sigma) := \smECE_{\sigma} - \sigma$. } \label{alg:smECE}
\end{algorithm}


Let us now discuss computation of $\smECE_{\sigma}$ for a given scale $\sigma$. To this end, we discretize the interval $[0,1]$, splitting it into $M$ equal length sub-intervals. For a sequence of observations $(f_i, y_i)$ we round each $r_i$ to the nearest integer multiple of $1/M$, mapping it to a bucket $b_i = \mathrm{round}(M f_i)$. In each bucket $b \in \{0, \ldots M\}$, we collect the residues of all observation falling in this bucket $h_b := \sum_{i : b_i = b} (f_{i} - y_i)$.

In the next step, we apply the~\Cref{clm:reflected-kernel-wrapping}, and produce a wrapping $\tilde{h}$ of the sequence $h$ --- extending it to integer multiples of $1/M$ in the interval $[-T, T+1]$ by pulling back $h$ through the map $\pi_R$.
The method $\smECEc_\sigma$ then proceeds to compute convolution $\tilde{h} \ast K$ with the discretization of the Gaussian kernel probability density function, i.e. $\tilde{K}_t := \exp(-t^2/2\sigma^2)$, and $K_t := \tilde{K}_t \left / \sum_i \tilde{K}_t \right.$.

This convolution $\tilde{r} := h \ast K$ can be computed in time $\Oh(Q \log Q)$, where $Q = MT$, using a Fast Fourier Transform, and is implemented in standard mathematical libraries. Finally, we report the sum of absolute values of the residuals $\sum |\tilde{r}_i|$ as an approximation to $\smECE_\sigma$ as an approximation to $\smECE_\sigma$.

\section{Discussion: Design Choices\label{sec:reliability-diagram}}

Here we discuss the motivation behind several design choices that may a-priori seem ad-hoc.
In particular, the choice to smooth the \emph{residuals} $(y_i - f_i)$ when computing
the $\smECE$, but to smooth the outcomes $y_i$ directly when plotting the reliability diagram.

For the purpose of constructing the reliability diagram, it might be tempting to plot a function $y'(f) := \hat{r}(f) + f$ (of smoothed residual as defined in~\eqref{eq:r_hat}, shifted back by the prediction $f$), as well as the smoothed density $\hat{\delta}(t)$, as in the definition of $\smECEc$. This is a fairly reasonable approach, unfortunately it has a particularly undesirable feature --- there is no reason for $y'(t) := \hat{r}(t) + t$ to be bounded in the interval $[0,1]$. It is therefore visually quite counter-intuitive, as the plot of $y(t)$ is supposed to be related with our guess on the average outcome $y$ given (slightly noisy version of) the prediction $t$.

As discussed in~\Cref{sec:method}, we instead consider the kernel regression on $y$, as opposed to the kernel regression on the residual $y-f$, and plot exactly this, together with the density $\hat{\delta}$. Specifically, let us define
\begin{equation}
\label{eq:y_hat}
    \hat{y}_{\cD,K}(t) := \frac{\E_{f,y\sim\cD} K(t, f) y}{\E_{f,y\sim \cD} K(t,f)}.
\end{equation}
and chose as the reliability diagram a plot of a pair of functions $t \mapsto \hat{y}_{\cD, K}(t)$ and $t \mapsto \hat{\delta}_{\cD, K}(t)$ --- the first plot is our estimation (based on the kernel regression) of the outcome $y$ for a given prediction $t$, the other is the estimation of the density of prediction $t$. As discussed in the \Cref{sec:method}, we will focus specifically on the kernel $K$ being the reflected Gaussian kernel, defined by \eqref{eq:reflected-gaussian}.

It is now tempting to define the calibration error related with this diagram, as an $\ECE$ of this new random variable over $[0,1] \times \{0,1\}$, analogously to the definition of $\smECE$, by considering 
\begin{equation}
    \smECEp_\sigma(\cD) := \int |\hat{y}_{\cD, K}(t) - t| \hat{\delta}_{\cD, K}(t) \dt.
    \label{eq:wrong-smece}
\end{equation}
This definition can be readily interpreted: for a random pair $(f,y)$ and an $\eta \sim \mathcal{N}(0, \sigma)$ independent, we can consider a pair $(f + \eta, y)$. It turns out that 
\begin{equation*}
    \smECEp_{\sigma}(\cD) = \ECE(\pi_R(f + \eta), y),
\end{equation*}
where $\pi_R : \R \to [0,1]$ collapses points that differ by reflection around integers (see~\Cref{sec:method}).

Unfortunately, despite being directly connected with more desirable reliability diagrams, and having more immediate interpretation as a $\ECE$ of a noisy prediction, this newly introduced measure $\smECEp$ has its own problems, and is generally mathematically much poorer-behaved than $\smECE$. In particular it is no longer the case that if we start with the perfectly calibrated distribution, and apply some smoothing with relatively large bandwidth $\sigma$, the value of the integral~\eqref{eq:wrong-smece} stays small. In fact it might be growing as we add more smoothing\footnote{This can be easily seen, if we consider the trivial perfectly calibrated distribution, where outcome $y \sim \Bernoulli(1/2)$ and prediction $f$ is deterministic $1/2$. Then $\smECEp_\sigma(\cD) = C\sigma$ for some constant $C \approx 0.79$.}.

Nevertheless, if we chose the correct bandwidth $\sigma^*$, as guided by the $\smECE$ consideration, the integral \eqref{eq:wrong-smece}, which is visually encoded by the reliability diagram we propose, should still be within constant factor from the actual $\smECE_\sigma^{*}(\cD)$, and hence provides a consistent calibration measure

\begin{lemma}
\label{lem:smece_poor}
    For any $\sigma$ we have
    \begin{equation*}
        \smECEp_{\sigma}(\cD) = \smECE_{\sigma}(\cD) \pm c \sigma,
    \end{equation*}
    where $c = \sqrt{2/\pi} \leq 0.8$.
    
    In particular, for $\sigma^*$ s.t. $\smECE_{\sigma^*}(\cD) = \sigma^*$, we have
    \begin{equation*}
        \smECEp_{\sigma^*}(\cD) \approx \smECE_{\sigma^*}(\cD).
    \end{equation*}
\end{lemma}
(The proof can be found in \Cref{sec:smece-properties}).


%% file: general-duality.tex
\section{General Metrics}
\label{sec:gen-metrics}
Our previous discussion implicitly assumed the 
the trivial metric on the interval $d(u,v) = |u-v|$.
We will now explore which aspects of our results extend to more general metrics over the interval $[0,1]$.
This is relevant if, for example, our application downstream of the predictor is more sensitive
to miscalibration near the boundaries.

The study of consistency measures with respect to general metrics is also motivated by the results of \cite{post-processing}. There it was shown that for any proper loss function $l$, there was an associated metric $d_{l}$ on $[0, 1]$ such that the predictor has small weak calibration error with respect to $d_l$ if and only if the loss $l$ cannot be significantly improved by post-composition with a Lipschitz function with respect to $d_l$. Specifically, they proved
\begin{equation*}
    \smce_{d_l}(\cD) \lesssim \E_{(f,y) \sim \cD}[l(f, y)] - \inf_{\kappa} \E_{(f, y) \sim \cD} [l(\kappa(f), y)] \lesssim \sqrt{\smce_{d_l}(\cD)},
\end{equation*}
where $\kappa : [0,1] \to [0,1]$ ranges over all functions Lipschitz with respect to the $d_l$ metric,
and $\smce_{d_l}(\cD)$ is the weak calibration error (as introduced by \citet{kakadeF08}, and extended
to general metrics in \citet{post-processing}, see~\Cref{def:wce}).

The most intuitive special case of the above result is the square loss function, which corresponds to a trivial metric on the interval $d(u,v) = |u-v|$.
In practice, different proper loss functions are also extensively used --- the prime example being the \emph{cross entropy loss} $l(f, y) := - (y \ln p + (1-y) \ln (1-p))$, which is connected with the metric $d_{logit}(u,v) := |\log(u/(1-u)) - \log(v/(1-v))|$ on $[0,1]$. 
Thus, we may want to generalize our results to also apply to non-trivial metrics.
%

\subsection{General Duality}
\label{sec:general-duality}
We will prove a more general statement of the duality theorem in \cite{UTC1}. Specifically, they showed that the minimization problem in the definition of the $\ldce$, can be dualy expressed as a maximal correlation between residual $r := y-f$ and a bounded Lipschitz function of the prediction $f$. This notion, which we will refer to as \emph{weak calibration error} first appeared in \cite{kakadeF08}, and was further explored in \cite{gopalan2022low,UTC1,post-processing}\footnote{
Weak calibration was called \emph{smooth calibration error} in \citet{gopalan2022low,post-processing}.
We revert back to the original terminology \emph{weak calibration error} to avoid confusion with the notion of $\smECE$  developed in this paper.}.

\begin{definition}[Weak calibration error]
\label{def:wce}
    For a distribution $\cD$ over $[0,1] \times \{0,1\}$ of pairs of prediction and outcome, and a metric $d$ on the space $[0,1]$ of all possible predictions, we define
    \begin{equation}
        \smce_{d}(\cD) := \sup_{w \in \mathcal{L}_{d}} \E_{f,y \sim \cD} (f-y) w(f),
    \end{equation}
    where the supremum is taken over all functions $w : [0,1] \to [-1,1]$ which are $1$-Lipschitz  with respect to the metric~$d$.\footnote{In \cite{UTC1}}
\end{definition}

For the trivial metric on the interval $d(u,v) = |u-v|$, $\smce$ was known to be linearly related with $\ldce$
by \cite{UTC1}.
We show in this paper that the duality theorem connecting $\smce$ and $\ldce$ holds much more generally, for a broad family of metrics.
    \begin{theorem}[\cite{UTC1}]
    \label{thm:duality}
        If a metric $d$ on the interval satisfies $d(u,v) \gtrsim |v - u|$ then $\smce_d \approx \ldce_d$.
    \end{theorem}
The more general formulation provided by~\Cref{thm:duality} can be shown by following closely the original proof step by step. We provide an alternate proof (simplified and streamlined) in \Cref{app:duality}.



\subsection{The $\ldce_{d_{logit}}$ is a consistent calibration measure with respect to $\ell_1$}

As in turns out, for a relatively wide family of metrics on the space of predictions (including the $d_{logit}$ metric), the associated calibration measures are consistent calibration measures \emph{with respect to the $\ell_1$ metric}. The main theorem we prove in this section is the following.

    \begin{theorem}
    \label{thm:consistent-ldce-d}
        If a metric $d : [0,1]^2 \to \mathbb{R} \cup \{\pm \infty\}$ satisfies $d(u,v) \gtrsim |u-v|$ and moreover for some $c > 0$,    \begin{equation*}
        \forall \varepsilon, \forall u, v \in [\varepsilon, 1-\varepsilon], \quad d(u,v) \leq |u - v| \varepsilon^{-c},
        \end{equation*}
        then $\ldce_{d}$ is a consistent calibration measure.
    \end{theorem}
The proof of this theorem (as is the case for many proofs of consistency for calibration measures) heavily uses the duality \Cref{thm:duality} --- since proving that a function is a consistent calibration measure amounts to providing a lower and upper bound, it is often convenient to use the $\ldce$ formulation for one bound and $\smce$ for the other.

The lower bound in \Cref{thm:consistent-ldce-d} is immediate --- since $d(u,v) \geq \ell_1(u,v)$, the induced Wasserstein distances on the space $[0,1] \times \{0,1\}$ satisfy the same inequality, hence $\ldce_{d} \geq \ldce_{\ell_1}$, and $\ldce_{\ell_1} \geq \ldce/2$ by~Claim~\ref{clm:definitions-equivalent}.

As it turns out, if the metric of interest is well-behaved except near the endpoints of the unit interval, we can also prove the converse inequality, and lower bound $\smce(\cD)$ by polynomial of $\smce_{d}(\cD)$.

\begin{lemma}
\label{lem:smce-lb}
    Let $d : [0,1]^2 \to \R_+ \cup \{\infty\}$ be any metric satisfying for some $c > 0$,    \begin{equation*}
        \forall \varepsilon, \forall u, v \in [\varepsilon, 1-\varepsilon], \quad d(u,v) \leq |u - v| \varepsilon^{-c},
        \end{equation*}
    then $\smce_d(\cD)^{q} \lesssim  \smce(\cD)$, where $q := \max(c+1, 2)$.
\end{lemma}
(Proof in~\Cref{app:smce-lb}.)

We are ready now to prove the main theorem here

\begin{proof}[Proof of \Cref{thm:consistent-ldce-d}]
    We have $\ldce_{d}(\cD) \geq \ldce_(\cD)/2$ by our previous discussion, on the other hand \Cref{thm:duality} and \Cref{lem:smce-lb} imply the converse inequality: \begin{equation*}
    \ldce_d(\cD) \approx \smce_d(\cD) \leq  \smce(\cD)^{1/q} \approx \ldce(\cD)^{1/q}.
    \end{equation*}
\end{proof}

\begin{corollary}
    For a metric induced by cross-entropy loss function $d_{logit}(u,v) := |\ln(u / (1-v)) - \ln(v / (1-v))|$, the $\smce_{d_{logit}}$ is a consistent calibration measure.
\end{corollary}
\begin{proof}
To verify the conditions of \Cref{thm:consistent-ldce-d} it is enough to check that $\logit(v) := \ln (v / (1-v))$ satisfies $\min(t, 1-t)^c \leq \frac{\mathrm{d}}{\dt} \logit (t) \leq C$. Since $\frac{\mathrm{d}}{\dt} \logit(t) = \frac{1}{t(1-t)}$, these conditions are satisfied with $c=1$ and $C=4$.
\end{proof}

\subsection{Generalized SmoothECE}
We now generalize the definition of SmoothECE to other metrics, and show that it remains
a consistent calibration measure with respect to its metric.
Motivated by the logit example discussed above, a concrete way to introduce a non-trivial metric on a space of predictions $[0,1]$, is to consider a continuous and increasing function $h : [0,1] \to \R \cup \{\pm\infty\}$, and the metric obtained by pulling back the metric from $\R$ to $[0,1]$ through $h$, i.e. $d_h(u, v) := |h(u) - h(v)|$.

Using the isomorphism $h$ between $([0,1], d_h)$ and a subinterval of $(\R \cup \{\pm \infty\}, |\cdot|)$, we can introduce a generalization of the notion of $\smECE$, where the kernel-smoothing is being applied in the image of $h$.

More concretely, and by analogy with  \eqref{eq:r_hat} and \eqref{eq:delta_hat}, for a probability distribution $\cD$ over $[0,1]\times \{0,1\}$, a kernel $K : \R \times \R \to \R_+$ and an increasing continuous map $h : [0,1] \to \R \cup \{\pm\infty\}$ we define
\begin{align*}
    \hat{r}_{K, h}(t) & := \frac{\E_{(f,y)} K(t, h(f)) (f-y)}{\E_{(f,y)} K(t, h(f))} \\
    \hat{\delta}_{K, h}(t) & := \E_{(f,y)} K(t, h(f)).
\end{align*}
Again, we define
\begin{equation*}
    \smECE_{K, h}(\cD) := \int \hat{r}_{K, h}(t) \hat{\delta}_{K, h}(t) \dt,
\end{equation*}
which simplifies to 
\begin{equation*}
    \smECE_{K, h}(\cD) = \int \left|\E_{(f,y)\sim \cD} K(t, h(f)) (f-y)\right| \dt.
\end{equation*}

As it turns out, with the duality theorem in place~(\Cref{thm:duality}) the entire content of \Cref{sec:smooth-ece} can be carried over in this more general context without much trouble.

Specifically, if we define $\smECE_{\sigma, h} := \smECE_{K_{N, \sigma}, h}$, where $K_{N, \sigma}$ is a Gaussian kernel with scale $\sigma$, then $\sigma \mapsto \smECE_{\sigma, h}(f,y)$ is non-increasing in $\sigma$, and therefore there is a unique fixed point $\sigma_*$ s.t. $\sigma_* = \smECE_{\sigma_*, h}(f,y)$. 

We can now define $\smECE_{*, h}(f,y) := \sigma_*$, and we have the following generalization of \Cref{thm:consistent-intro}, showing that SmoothECE remains a consistent calibration even under different metrics.
\begin{theorem}
\label{thm:consistent-smece-general}
For any increasing and continuous function $h : [0,1] \to \mathbb{R} \cup \{\pm \infty\}$, if we define $d_h : [0,1]^2 \to \mathbb{R}_+$ to be the metric $d_h(u,v) = \max(|h(u) - h(v)|, 2)$ then
\begin{equation*}
    \ldce_{d_h}(\cD) \lesssim \smECE_{*, h}(\cD) \lesssim \sqrt{\ldce_{d_h}(\cD)}.
\end{equation*}
\end{theorem}
(Proof in \Cref{sec:proof-of-consistent-smece-general}.)

Note that if the function $h$ is such that the associated metric $d_h$ satisfies~the conditions of~\Cref{thm:consistent-ldce-d}, as an additional corollary we can deduce that $\smECE_{*, h}$ is also a consistent calibration measure in a standard sense.

\subsection{Obtaining perfectly calibrated predictor  via post-processing}
One of the appealing properties of the notion $\ldce$ as it was introduced in \cite{UTC1}, was the theorem stating that if a predictor $(f, y)$ is close to calibrated, then in fact a nearby perfectly calibrated predictor can be obtained simply by post-processing all the predictions by a univariate function. Specifically, they showed that for a distribution $\cD$ over $[0,1] \times \{0,1\}$, there is $\kappa : [0,1] \to [0,1]$ such that for $(f,y) \sim \cD$ the pair $(\kappa(f), y)$ is perfectly calibrated and moreover $\E |\kappa(f) - f| \lesssim \sqrt{\ldce(\cD)}$.

As it turns out, through the notion of $\smECE_h$ we can prove a similar in spirit statement regarding the more general distances to calibration $\ldce_{d_h}$.
The only difference is that we allow the post-processing $\kappa$ to be a randomized function.
\begin{theorem}
\label{thm:udce}
    For any increasing function $h : [0,1] \to \mathbb{R} \cup \{\pm\infty\}$, and any distribution $\cD$ supported on $[0,1] \times \{0, 1\}$, there is a \emph{probabilistic} function $\kappa : [0,1] \to [0,1]$ such that for $(f,y) \sim \cD$, the pair $(\kappa(f), y)$ is perfectly calibrated and
    \begin{equation*}
        \E d_h(\kappa(f), f) \lesssim \smECE_{*,h}(\cD),
    \end{equation*}
    where $d_h$ is the metric induced by $h$. In particular
    \begin{equation*}
        \E d_h(\kappa(f), f) \lesssim \sqrt {\ldce_{d_h}(\cD)}.
    \end{equation*}
\end{theorem}
Proof in the appendix.

%% file: experiments.tex
\section{Experiments}
\label{sec:experiments}


We include several experiments demonstrating our method 
on public datasets in various domains, from deep learning to meteorology.
The sample sizes vary between several hundred to 50K, to show how our method behaves for different data sizes.
In each setting, we compare the classical binned reliability diagram
to the smooth diagram generated by our Python package.
Our diagrams include
bootstrapped uncertainty intervals for the SmoothECE,
as well as kernel density estimates of the predictions
(at the same kernel bandwidth $\sigma^*$ used to compute the SmoothECE).
For binned diagrams, the number of bins is chosen to be optimal for the regression test MSE loss,
optimized via cross-validation.
Code to reproduce these figures is available at \url{https://github.com/apple/ml-calibration}.

\begin{figure}[p]
     \centering
     \includegraphics[width=0.8\textwidth]{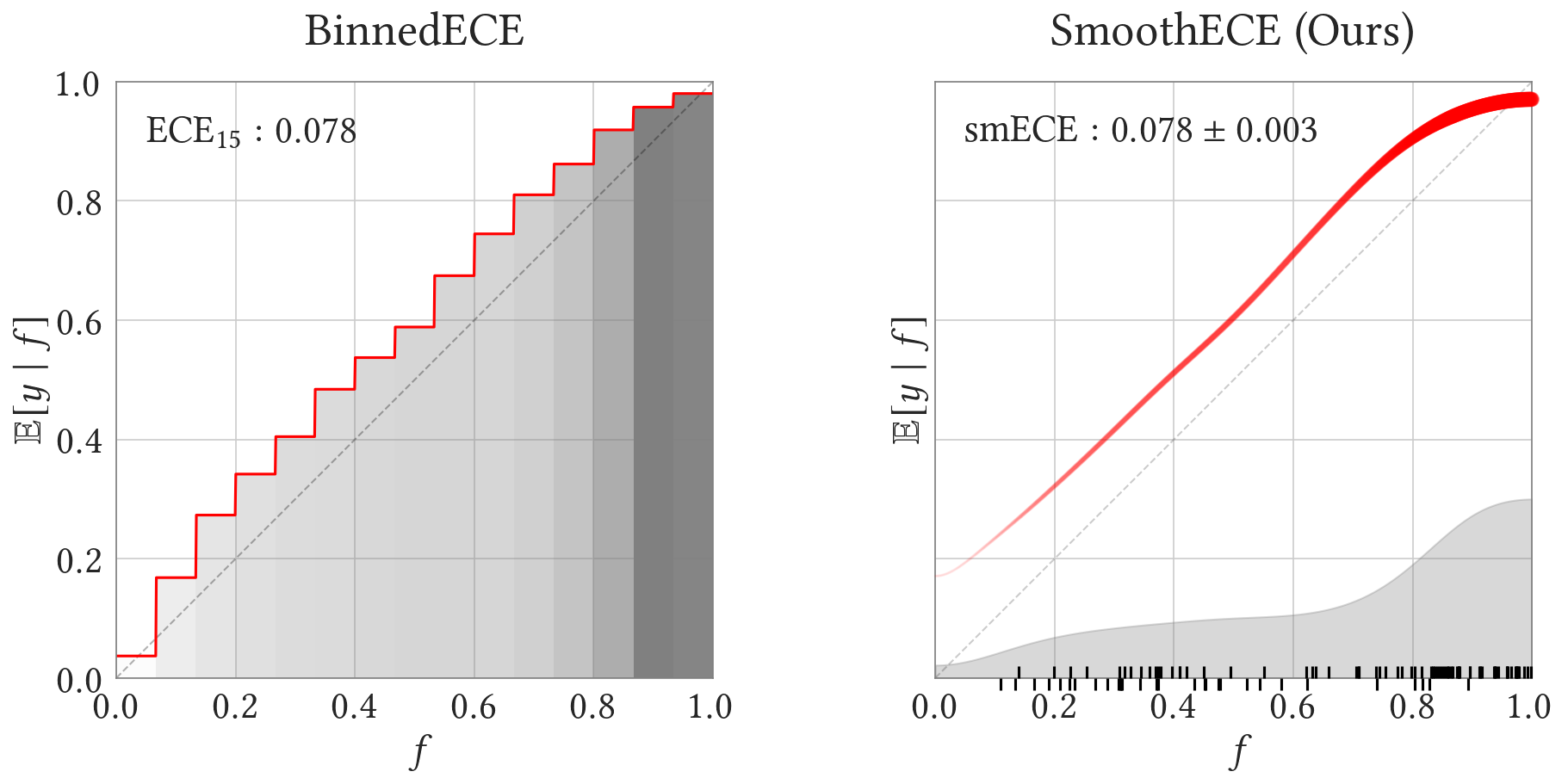}
     \caption{Confidence calibration of ResNet34 on ImageNet. Data from \citet{hollance}.}
     \label{fig:resnet}
\end{figure}

\paragraph{Deep Networks.}
Figure~\ref{fig:resnet} shows the confidence calibration of ResNet32 \citep{he2016deep} on the ImageNet validation set \citep{deng2009imagenet}. 
ImageNet is an image classification task with 1000 classes, and has a validation set of 
50,000 samples.
In this multi-class setting, the model $f$ outputs a distribution over $k=1000$ classes, $f: \cX \to \Delta_k$.
Confidence calibration is defined
as calibration of the pairs $( \argmax_{c \in [k]} f_c(x) ~~,~~  \1\{f(x) = y\})$,
which is a distribution over $[0, 1] \x \{0, 1\}$.
That is, confidence calibration measures the agreement between confidence 
and correctness of the predictions.
We use the publicly available data from \citet{hollance}, evaluating the models trained by \citet{rw2019timm}.

\begin{figure}[p]
     \centering
     \includegraphics[width=0.8\textwidth]{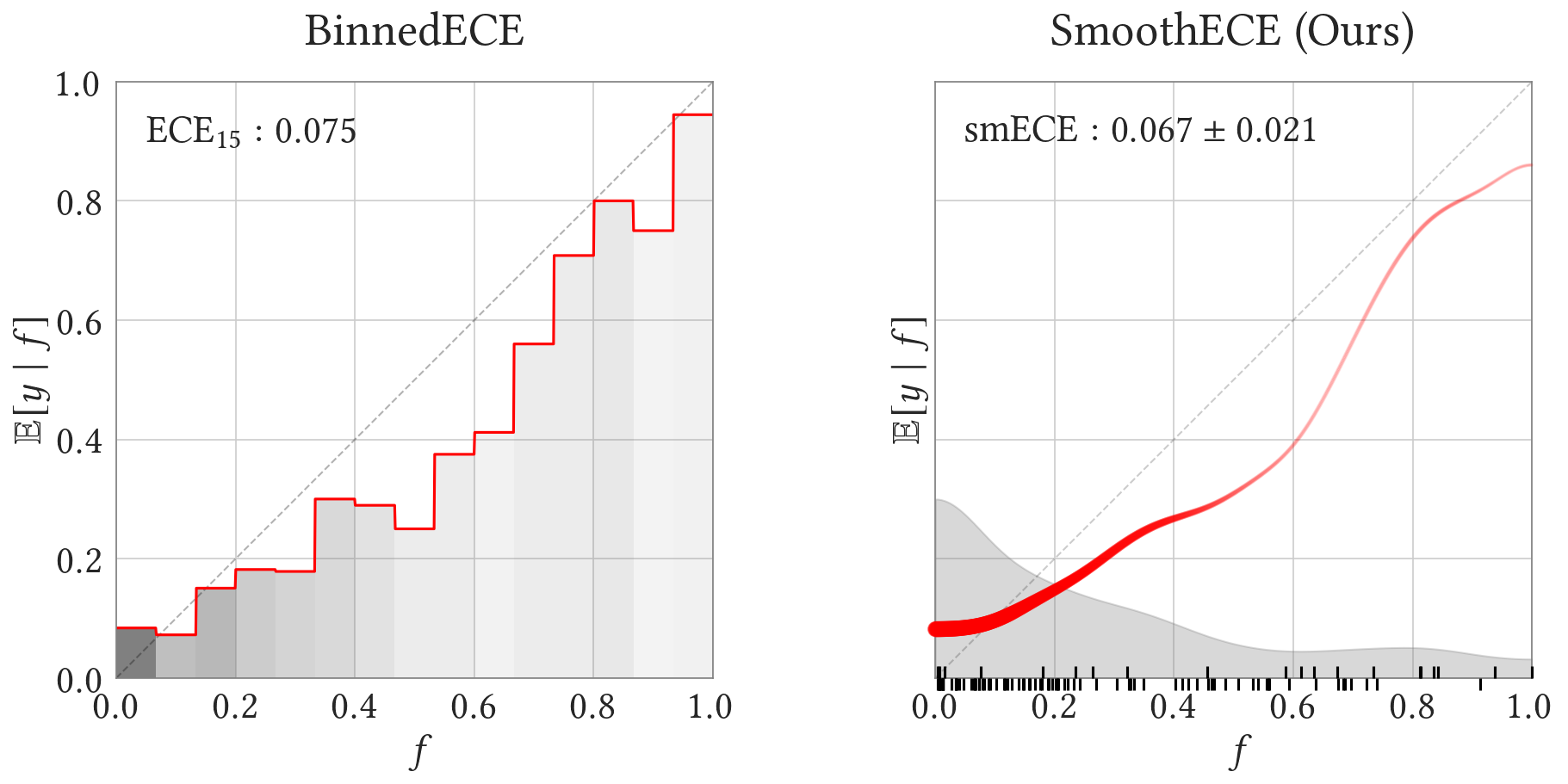}
     \caption{Calibration of solar flare forecasts over a 731 day period.
     Data from \citet{leka2019comparison,dimitriadis2023evaluating}.
     }
     \label{fig:solar}
\end{figure}

\paragraph{Solar Flares.}
Figure~\ref{fig:solar} shows the calibration of a method for forecasting solar flares, over a period of 731 days.
We use the data from \citet{leka2019comparison},
which was used to compare reliability diagrams 
in \citet{dimitriadis2021stable}.
Specifically, we consider forecasts of the event that a class C1.0+ solar flare occurs on a given day,
made by the DAFFS forecasting model developed by NorthWest Research Associates.
Overall, such solar flares occur on 25.7\% of the 731 recorded days.
We use the preprocesssed data from the replication code
at: \url{https://github.com/TimoDimi/replication_DGJ20}.
For further details of this dataset, we refer the reader to \citet[Section 6.1]{dimitriadis2023evaluating} and \citet{leka2019comparison}.

\begin{figure}[p]
     \centering
     \includegraphics[width=0.8\textwidth]{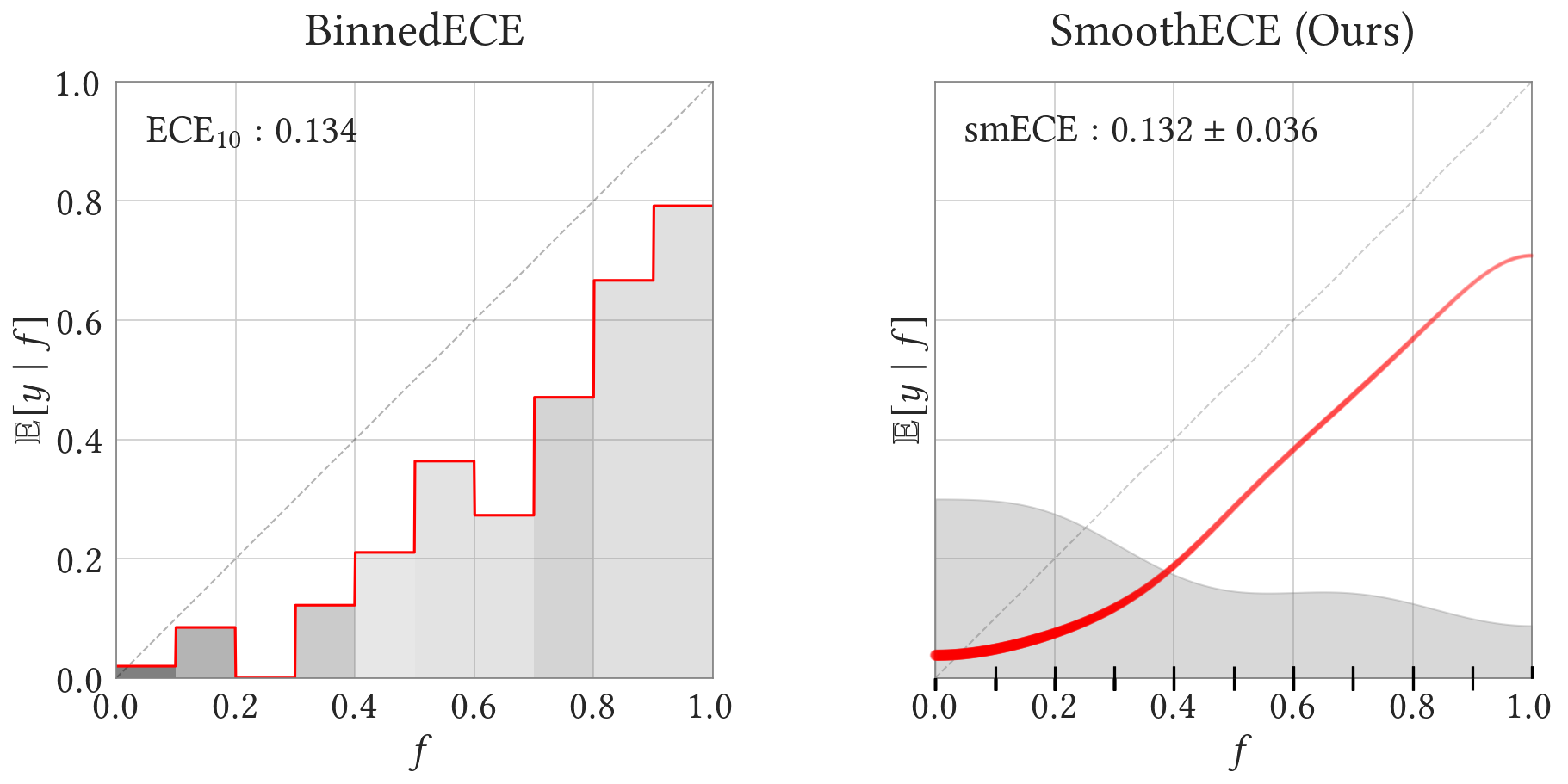}
     \caption{Calibration of daily rain forecasts in Finland in 2003.
     Data form \citet{FMI}.}
     \label{fig:rain}
\end{figure}

\paragraph{Precipitation in Finland.}
Figure~\ref{fig:rain} shows the calibration of 
daily rain forecasts 
made by the Finnish Meteorological Institute (FMI) in 2003,
for the city of Tampere in Finland.
Forecasts are made for the probability that precipitation exceeds $0.2$mm over a 24 hour period;
the dataset includes records for 346 days \citep{FMI}.

\begin{figure}[p]
     \centering
     \includegraphics[width=0.8\textwidth]{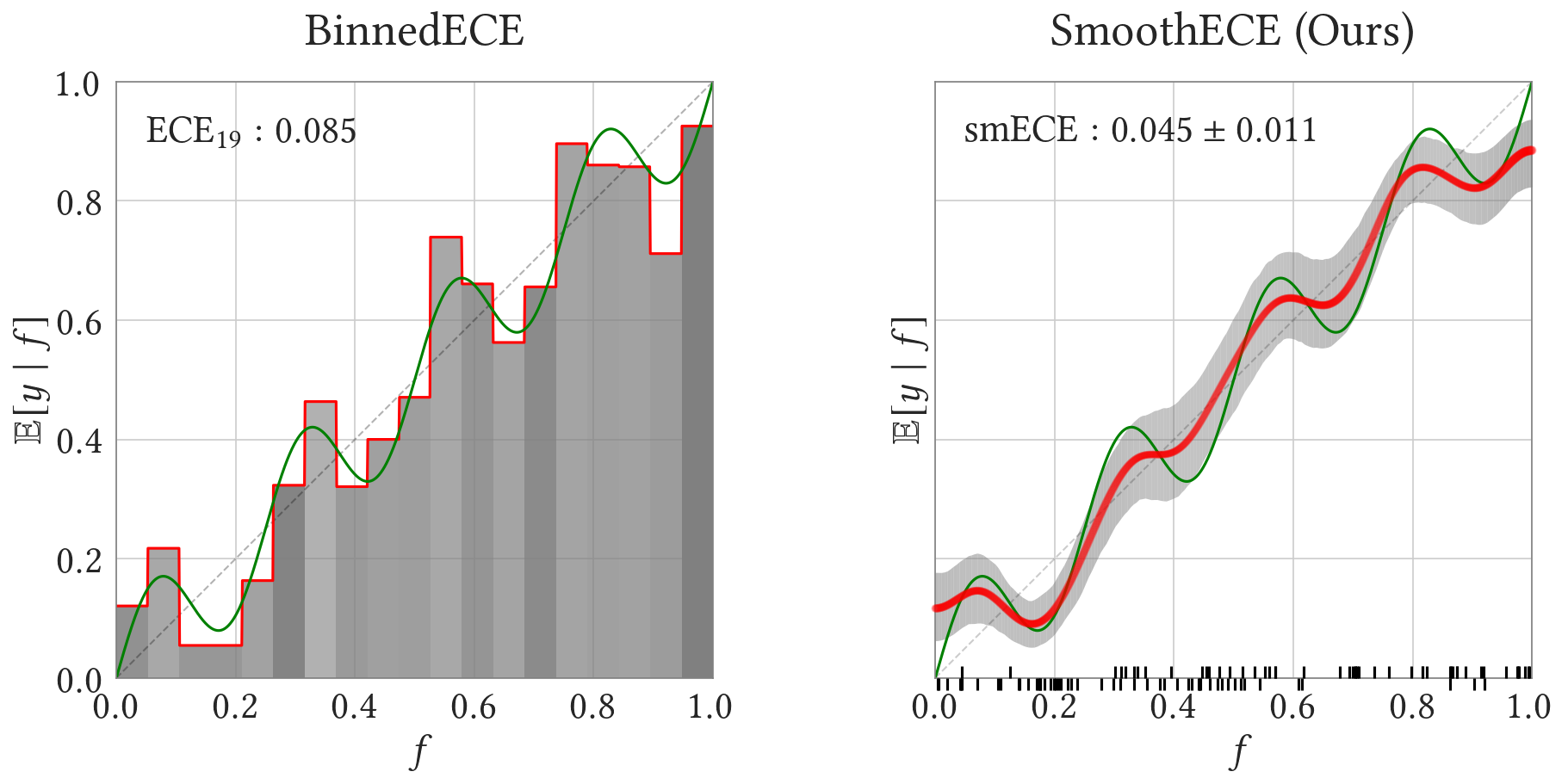}
     \caption{Calibration of synthetic data in a toy example.
     Here, instead of kernel density estimates, we show bootstrapped uncertainity bands around our estimated
     regression function.}
     \label{fig:toy}
\end{figure}

\paragraph{Synthetic Data.}
For demonstration purposes, we apply our method to a simple synthetic dataset in Figure~\ref{fig:toy}.
One thousand samples $f_i \in [0, 1]$ are drawn uniformly in the interval $[0, 1]$,
and the conditional distribution of labels $\E[ y_i \mid f_i ]$ is given by the green line
in Figure~\ref{fig:toy}. Note that the true conditional distribution is non-monotonic in this example.

\paragraph{Limitations.}
One limitation of our method is that since it is generic, there 
may be better tools to use in special cases, when we can assume more structure
in the prediction distributions.
For example, if the predictor is known to only output a small finite set of possible
probabilities, then it is reasonable to simple estimate conditional probabilities
by using these points as individual bins.
The rain forecasts in Figure~\ref{fig:rain} have this structure, since the
forecasters only predict probabilities in multiples of 10\% -- in such cases,
using bins which are correctly aligned is a very reasonable option.
Finally, note that the boostrapped uncertainty bands shown in our reliability diagrams
should not be interpreted as confidence intervals for the \emph{true} regression function.
Rather, the bands reflect the sensitivity of our particular regressor under resampling the data.

%% file: conclusion.tex
\section{Conclusion}
\label{sec:conclusion}

We have presented a method of computing calibration error
which is both mathematically well-behaved
(i.e. \emph{consistent} in the sense of \citet{UTC1}),
and can be visually represented in a reliability diagram.
We also released a python package which efficiently implements our suggested method.
We hope this work aids practitioners in computing, analyzing,
and visualizing the reliability of probabilistic predictors.

%% file: math.tex
\section{Appendix\label{app:proofs}}

\subsection{Proof of \Cref{thm:consistent-intro}}

In this section we will prove \Cref{lem:lb} and \Cref{lem:ub}, two main steps in the proof of \Cref{thm:consistent-intro}, corresponding to respectively lower and upper bound. As it turns out, those two lemmas are true for a much wider class of kernels. The restriction on the kernel $K$ to be a Gaussian kernel stems from the monotonicity property (\Cref{lem:monotninicity}), which was convenient for us to define the scale invariant measure $\smECE_*$ by considering a fix-point scale $\sigma^*$. In~\Cref{sec:reflected-gaussian-properties} we will show that the Reflected Gaussian kernel satisfies the conditions of \Cref{lem:lb} and \Cref{lem:ub}.

We will first define a dual variant of $\ldce$.
\begin{definition}
    We define the weak calibration error to be the maximal correlation of the residual $(f-y)$ with a $1$-Lipschitz function and $[-1,1]$ bounded function of a predictor, i.e.
    \begin{equation*}
        \smce(\cD) := \sup_{w \in \mathcal{L}} \E_{(f,y) \sim \cD} w(f) (f-y),
    \end{equation*}
    where $\mathcal{L}$ is a family of all $1$-Lipschitz functions from $[0,1]$ to $[-1, 1]$.
\end{definition}
To show that $\smECE^*$ is a consistent calibration measure we will heavily use the duality theorem proved in \cite{UTC1} --- the $\smce$ and $\ldce$ are (up to a constant factor) equivalent. A similar statement is proved in this paper, in a greater generality (see~\Cref{thm:duality}).
\begin{theorem}[\cite{UTC1}]
\label{thm:duality-utc1}
    For any distribution $\cD$  over $[0,1] \times \{0, 1\}$ we have
    \begin{equation*}
        \ldce(\cD) \leq \smce(\cD) \leq 2 \ldce(\cD).
    \end{equation*}
\end{theorem}
Intuitively, this is useful since showing that a new measure $\smECE$ is a consistent calibration measure corresponds to upper and lower bounding it by polynomials of $\ldce$. With the duality theorem above, we can use the minimization formulation $\ldce$ for one direction of the inequality, and the maximization formulation $\smce$ for the other.

Indeed, we will first show that $\smce$ is upper bounded by $\smECE$ if we add the penalty parameter for the ``scale'' of the kernel $K$.
\begin{lemma}
\label{lem:lb}
Let $U\subset \R$ be (possible infinite) interval containing $[0,1]$ and $K : U \times U \to \R$ be a non-negative symmetric kernel satisfying for every $t_0 \in [0,1]$, $\int K(t_0, t) \dt = 1$, and $\int |t - t_0| K(t, t_0) \dt \leq \gamma$.  Then
\begin{equation*}
    \smce(\cD) \leq \smECE_{K}(\cD) + \gamma.
\end{equation*}
\end{lemma}
\begin{proof}
    Let us consider an arbitrary 1-Lipschitz function $w : [0,1] \to [-1, 1]$, and take $\eta \sim K$ as in the lemma statement. Since kernel $K$ is nonnegative, and $\int K(t, t_0) \dt = 0$, we can sample triple $(\tilde{f}, f, y)$ s.t. $(f,y) \sim \cD$, and $\tilde{f}$ is distributed according to density $K(\cdot, f)$. In particular $\E |\tilde{f} - f| \leq \gamma$.

    We can bound now
    \begin{align}
        \E_{(f,y) \sim \cD} [w(f) (f-y)] & \leq \E [w(\tilde{f}) (f - y)] + \E |f - \tilde{f}| |f-y| \nonumber \\
        & \leq \gamma + \E \left[w(\tilde{f}) (f - y) \right]  \label{eq:lem-lb-split}.
    \end{align}

    We now observe that
    \begin{equation*}
        \E[(f-y) | \tilde{f} = t] = \frac{\E_{f,y} K(t,f)(f-y)}{\E_{f,y} K(t,f)} = \hat{r}(t),
    \end{equation*}
    and the marginal density of $\tilde{f}$ is exactly
    \begin{equation*}
        \mu_{\tilde{f}}(t) = \E_{(f,y) \sim \cD} K(t,f) = \hat{\delta}(t).
    \end{equation*}
    
    This leads to
    \begin{equation}
        \E\left[ w(\tilde{f}) (f-y) \right] = \int w(t)\hat{r}(t) \hat{\delta}(t) \dt \leq \int |\hat{r}(t)| \hat{\delta}(t) \dt = \smECE_K(f,y).
        \label{eq:lem-lb-smece}
    \end{equation}
    Combining \eqref{eq:lem-lb-split} and \eqref{eq:lem-lb-smece} we conclude the statement of this lemma.
\end{proof}

To show that $\smECE_K(\cD)$ is upper bounded by $\ldce$, we will first show that $\smECE_K$ is zero for perfectly calibrated distributions, and then we will show that for well-behaved kernels $\smECE_K(\cD)$ is Lipschitz with respect the Wasserstein distance on the space of distributions.

\begin{claim}
\label{clm:perfectly-calibrated}
    For any perfectly calibrated distribution $\cD$ and for any kernel $K$ we have
    \begin{equation*}
        \smECE_K(\cD) = 0.
    \end{equation*}
\end{claim}
\begin{proof}
    Indeed, by the definition of $\hat{r}$ we have
    \begin{equation*}
        \hat{r}(t) = \frac{\E_{f,y} K(f,t) (f-y)}{\E_{f,y} K(f,t)},
    \end{equation*}
    Since the distribution $\cD$ is perfectly calibrated, we have $\E_{(f,y) \sim \cD} [ (f-y) | f] = 0$, hence 
    \begin{equation*}
        \E_{f,y} [K(f,t) (f-y)] = \E_{f}\left[\E_{(f,y) \sim \cD} [K(f,t) (f-y) | f]\right] = \E_f\left[ K(f,t) \E_{(f,y)\sim \cD} [ (f-y) | f]\right] = 0. 
    \end{equation*}
    This means that the function $\hat{r}(t)$ is identically zero, and therefore
    \begin{equation*}
        \smECE_K(\cD) = \int_t |\hat{r}(t)| \hat{\delta}(t) \dt = 0.
    \end{equation*}
\end{proof}

\begin{lemma}
\label{lem:ub}
Let $K$ be a symmetric, non-negative kernel, such that for and let $\lambda \leq 1$ be a constant such that 
for any $t_0, t_1 \in [0,1]$ we have $\int |K(t_0, t) - K(t_1, t)|\dt \leq |t_0 - t_1|/\lambda$. Let $\cD_1, \cD_2$ be a pair of distributions over $[0,1]\times \{0,1\}$. Then
\begin{equation*}
    |\smECE_K(\cD_1) - \smECE_K(\cD_2)| \leq \left(\frac{1}{\lambda} + 1\right) W_1(\cD_1, \cD_2),
\end{equation*}
where the Wasserstein distance is induced by the metric $d((f_1, y_1), (f_2, y_2)) = |f_1 - f_2| + |y_1 - y_2|$ on $[0,1] \times \{0,1\}$. 
\end{lemma}
\begin{proof}
We have
\begin{equation*}
    \smECE_K(\cD) = \int \left|\E_{(f,y) \sim \cD} [K(t, f) (y - f)] \right| \dt.
\end{equation*}
If we have a coupling $(f_1, f_2, y_1, y_2)$ s.t. $\E [|f_1 - f_2| + |y_1 - y_2|] \leq \delta$, $(f_1, y_1) \sim \cD_1$ and $(f_2, y_2) \sim \cD_2$, then by triangle inequality we can decompose
\begin{align*}
    |\smECE_K(\cD_1) - \smECE_K(\cD_2)| & \leq
    \int  \E_{(f_1, f_2,y_1, y_2)} [|K(t, f_1) - K(t, f_2)| |y_1 - f_1| \dt \\
    & + \int \E_{(f_1, f_2, y_1, y_2)} [K(t, f_2) (|f_1 - f_2| + |y_1 - y_2|] \dt.
\end{align*}
We can bound those two terms separately
\begin{equation*}
    \int \E_{(f_1, f_2, y_1)} [|K(t, f_1) - K(t, f_2)| |y_1 - f_1|] \dt \leq \E_{(f_1,f_2, y_1)} \int |K(t, f_1) - K(t, f_2)| \dt \leq \frac{1}{\lambda} \E[ |f_1 - f_2|] \leq \delta/\lambda,
\end{equation*}
and similarly
\begin{equation*}
    \int \E \left[K(t, f_2)  (|f_1 - f_2| + |y_1 - y_2|)\right]\dt = \E\left[ \int_t K(t, f_2) \dt  \cdot (|f_1 - f_2| + |y_1 - y_2|\right] = \E[|f_1 - f_2| + |y_1 - y_2|] \leq \delta.
\end{equation*}
\end{proof}
\begin{corollary}
    Under the same assumptions on $K$ as in \Cref{lem:ub}, for any distribution $\cD$ over $[0,1] \times \{0,1\}$,
    \begin{equation*}
        \smECE_K(\cD) \leq \left(\frac{1}{\lambda} + 1\right) \ldce(\cD).
    \end{equation*}
\end{corollary}
\begin{proof}
    By definition of the $\ldce$ there is a perfectly calibrated distribution $\cD'$, such that $W_{1}(\cD, \cD') \leq \ldce(\cD)$, since the $W_1(\cD, \cD')$ is decreasing as we change the underlying metric to a smaller one. By~\Cref{clm:perfectly-calibrated}, $\smECE_K(\cD') = 0$, and the corollary follows directly from~\Cref{lem:ub}.
\end{proof}

\subsection{Facts about reflected Gaussian kernel}
\label{sec:reflected-gaussian-properties}
We wish to now argue that \Cref{lem:lb} and \Cref{lem:ub} imply the more specialized statements \Cref{lem:lb-intro} and \Cref{lem:ub-intro} respectively --- the reflected Gaussian kernel $K_{N, \sigma}$ satisfies conditions of \Cref{lem:lb} and \Cref{lem:ub} with $\gamma$ and $\lambda$ proportional to $\sigma$. We 

\begin{lemma}
    Reflected Gaussian kernel $\tilde{K}_{N, \sigma}$ defined by \eqref{eq:reflected-gaussian} satisfies
    \begin{enumerate}
        \item For every $t_0$, we have $\int \tilde{K}_{N, \sigma}(t, t_0) \dt = 1.$
        \item For every $t_0$, we have $\int |t - t_0| \tilde{K}_{N, \sigma}(t, t_0) \dt \leq \sqrt{2/\pi} \sigma.$
        \item For every $t_0, t_1$, we have $\int |\tilde{K}_{N, \sigma}(t, t_0) - \tilde{K}_{N, \sigma}(t, t_0)| \dt \leq  |t_0 - t_1|/(2 \sigma).$
    \end{enumerate}
\end{lemma}
\begin{proof}
    For any given $t_0$, the function $\tilde{K}_{N, \sigma}(t_0, \cdot)$ is a probability density function of a random variable $\pi_R(t_0 + \eta)$ where $\eta \sim \mathcal{N}(0, \sigma)$ and $\pi_R : \R \to [0, 1]$ is defined in~\Cref{sec:method}. In particular, we have $|\pi_R(x) - \pi_R(y)| \leq |x-y|$.

    The property 1 is satisfied, since the $\tilde{K}_{N, \sigma}(\cdot, t_0)$ is a probability density function.

    The property 2 follows since
    \begin{align*}
        \int |t - t_0| \tilde{K}_{N, \sigma}(t, t_0) \dt & = \E_{\eta \sim \mathcal{N}(0, \sigma} |\pi_R(t_0 + \eta) - t_0| 
         = \E_{\eta \sim \mathcal{N}(0, \sigma} |\pi_R(t_0 + \eta) - \pi_R(t_0)| \\
         & \leq \E_{\eta \sim \mathcal{N}(0, \sigma} |\eta| = \sigma \sqrt{2/\pi}.
    \end{align*}
    Finally, the property 2 again follows from the same fact for a Gaussian random variable: the integral $|\tilde{K}_{N, \sigma}(t, t_0) - \tilde{K}_{N, \sigma}(t, t_0)|$ is just a total variation distance between $\pi_R(t_0 + \eta)$ and $\pi_R(t_1 + \eta)$ where $\eta \sim \mathcal{N}(0, \sigma)$, but by data processing inequality we have 
    \begin{equation*}
        TV(\pi_R(t_0 + \eta), \pi_R(t_1 + \eta)) \leq TV(t_0 + \eta, t_1 + \eta) \leq |t_0 - t_1| / (2 \sigma).
    \end{equation*}
    Where the last bound on the total variation distance between two one-dimension Gaussians is a special case of Theorem 1.3 in \cite{devroye2018total}\footnote{This special case, where the two variances are equal, is in fact an elementary calculation.}.
\end{proof}

\begin{definition}
\label{def:proper-kernel-family}
    We say that a paramterized family of kernels $K_{\sigma} : U \times U \to \R$ where $[0,1] \subset U \subset \R$ is a \emph{proper kernel family}
    if for any $\sigma_1 \leq \sigma_2$ there is a non-negative kernel $H_{\sigma_1, \sigma_2} : U \times U \to \R$, satisfying $\| H_{\sigma_1, \sigma_2}\|_{1 \to 1}\leq  1$ and $K_{\sigma_2} = K_{\sigma_1} \ast H_{\sigma_1, \sigma_2}$.
    
    Here the notation $K \ast H$ is denotes
    \begin{equation*}
        [K \ast H](t_1, t_2) := \int_U K(t_1, t) H(t, t_2) \dt,
    \end{equation*}
    and
    \begin{equation*}
        \|H\|_{1 \to 1} := \sup_{t_0 \in U} \int_{U} |H(t_0, t)|\dt.
    \end{equation*}
\end{definition}

\begin{claim}
    \label{clm:reflected-gaussian-proper}
    The family of reflected Gaussian kernels $\tilde{K}_{N, \sigma}$ is a proper kernel family, with
    \begin{equation*} \tilde{K}_{\sigma_1, N} = \tilde{K}_{\sigma_2, N} \ast \tilde{K}_{\sqrt{\sigma_1^2 - \sigma_2^2}, N}.
    \end{equation*}
\end{claim}
\begin{proof}
    Let $\sigma_3 := \sqrt{\sigma_1^2 - \sigma_2^2}$, 
    we wish to show that $\tilde{K}_{\sigma_1, N} = \tilde{K}_{\sigma_2, N} \ast \tilde{K}_{\sigma_3, N}$. In order to show this, it is enough to prove that for any $f$, we have $f \ast \tilde{K}_{\sigma_1, N} = f \ast \tilde{K}_{\sigma_2, N} \ast \tilde{K}_{\sigma_3, N}$. This is true by \Cref{clm:reflected-kernel-wrapping}, since this property holds for standard Gaussian kernel $K_{\sigma_2, N} \ast K_{\sigma_3, N} = K_{\sigma_1, N}$ (it is here equivalent to saying that for two independent random variables $Z_2 \sim \mathcal{N}(0, \sigma_2)$ and $Z_3 \sim \mathcal{N}(0, \sigma_3)$ we have $Z_2 + Z_2 \sim \mathcal{N}(0, \sigma_1))$.
\end{proof}

\subsection{Useful properties of $\smECE$.}
\label{sec:smece-properties}
\begin{lemma}[Monotonicity of $\smECE$]
\label{lem:monotninicity}
Let $K_{\sigma}$ be any proper kernel family parameterized by $\sigma$ (see~\Cref{def:proper-kernel-family}).  If $\sigma_1 \leq \sigma_2$, then
\begin{equation*}
\smECE_{K_{\sigma_1}}(\cD) \geq \smECE_{K_{\sigma_2}}(\cD).
\end{equation*}
\end{lemma}
\begin{proof}
    Let us define 
    \begin{equation*}
        h_{\sigma}(t) := \E_{(f, y)\sim \cD} K_{\sigma}(t, f) (f - y) = \hat{r}(t) \hat{\delta}(t),
    \end{equation*}
    such that 
    \begin{equation*}
        \smECE_{K_\sigma}(\cD) = \|h_{\sigma}\|_1 := \int |h_{\sigma}(t)| \dt.
    \end{equation*}
    Since $\sigma_1 \leq \sigma_2$ and $K_{\sigma}$ is a proper kernel family, we can write $K_{\sigma_2} = K_{\sigma_1} \ast H_{\sigma_1, \sigma_2}.$
    
    We have now,
    \begin{align*}
        h_{\sigma_1} \ast H_{\sigma_1, \sigma_2} & =  \left(\E_{(f,y)} (f-y) K_{\sigma_1}(\cdot, f)\right) \ast  H_{\sigma_1, \sigma_2} \\
        & = \E_{f,y} (f-y) [K_{\sigma_1} \ast H_{\sigma_1,\sigma_2}(\cdot, f)] = \E_{f-y} (f-y) K_{\sigma_2}(\cdot, f) \\
        & = h_{\sigma_2}.
    \end{align*}
    On the other hand for any function $f$ we have
    $\|f \ast H_{\sigma_1, \sigma_2}\|_1 \leq \|f\|_1 \|H_{\sigma_1, \sigma_2}\|_{1 \to 1}$, and $\|H_{\sigma_1, \sigma_2}\|_{1\to 1} \leq 1$ by the definition of proper kernel family. Therefore
    \begin{equation*}
    \end{equation*}
\end{proof}
\begin{corollary}
    In particular for $\sigma_1 \leq \sigma_2$ we have $\smECE_{\sigma_2}(\cD) \leq \smECE_{\sigma_1}(\cD)$.
\end{corollary}
\begin{proof}
Reflected Gaussian kernels form a proper kernel family by \Cref{clm:reflected-gaussian-proper}.
\end{proof}
\begin{lemma}
    For any $\sigma$, we have $\smECEp_{\sigma}(\cD) = \smECE_{\sigma}(\cD) \pm \sigma \sqrt{2/\pi}$. 
\end{lemma}
\begin{proof}
    Let 
    \begin{equation*}
        \hat{f}(t) := \frac{\E_{f,y} \tilde{K}_{N, \sigma}(t, f) f}{\E_{f,y} \tilde{K}_{N, \sigma}(t, f)}.
    \end{equation*}
    We have
    \begin{align*}
        |\smECEp_{\sigma}(f,y) - \smECE_{\sigma}(f,y)|
        & \leq \int |\hat{f}(t) - t| \hat{\delta}(t) \dt \\
        & \leq \int \E_f [\tilde{K}_{N, \sigma}(t, f) |f-t|] \dt \\
        & = \E_f \int K_{\sigma}(t, f) |f-t|\dt \\
        & = \E_{f} \E_{Z \sim \mathcal{N}(f, \sigma)} |f - \pi_R(Z)| \\
        & \leq \E_{Z \sim \mathcal{N}(0,\sigma)} |Z| = \sqrt{2 / \pi}.
    \end{align*}
\end{proof}

\subsection{Equivalence between definitions of $\ldce$ for trivial metric}

The $\ldce(\cD)$ was defined in \cite{UTC1} as a Wasserstein distance to the set of perfectly calibrated distributions over $X := [0,1] \times \{0,1\}$, where $X$ is equipped with a metric
    \begin{equation*}
        d_{1}( (f_1, y_1), (f_2, y_2)) := 
        \begin{cases}
            |f_1 - f_2| & \text{if } y_1 = y_2 \\
            \infty &\text{otherwise}
        \end{cases}.
    \end{equation*}
While generalizing the notion to that of $\ldce_d$, where $d$ is a general metric on $[0,1]$, we chose a different metric on $X$ (specifically, we put a different metric on the second coordinate), that is $\tilde{d}((f_1, y_1), (f_2, y_2)) = d(f_1, f_2) + |y_1 - y_2|$.

As it turns out, for the case of a trivial metric on the space of predictions, this choice is inconsequential, but the new definition has better generalization properties.
\begin{claim}
\label{clm:definitions-equivalent}
    For the metric $\ell_1(f_1, f_2) = |f_1 - f_2|$, we have
        $\ldce(\cD) \lesssim \ldce_{\ell_1}(\cD) \leq \ldce(\cD)$, for some universal constant $c$.
\end{claim}
\begin{proof}
The lower bound $\ldce_{\ell_1} \leq \ldce$ is immediate, since $\ldce_{\ell_1}$ is a distance of $\cD$ to $\mathcal{P}$ with respect to a Wasserstein distance induced by the metric $d_1$ on $[0,1] \times \{0, 1\}$, $\ldce$ is the Wasserstein distance with respect to the metric $d_2$, and we have a pointwise bound $d_1(u,v) \leq d_2(u,v)$, implying $W_{1, d_1}(\cD_1, \cD_2) \leq W_{1, d_2}(\cD_1, \cD_2)$.

The other bound follows from~\Cref{thm:duality} and~\Cref{thm:duality-utc1} --- $\ldce$ and $\ldce_{\ell_1}$ are within constant factor from $\smce_{\ell_1}$.
\end{proof}

\subsection{Proof of \Cref{lem:smce-lb} \label{app:smce-lb}}
\begin{proof}
    Let us take $w(x) : [0,1] \to [-1,1]$ as in the definition of $\smce_d$, a $1$-Lipschitz function with respect to the metric $d$, such that $\E (y-f) w(f) = \smce_d(f,y) = \varepsilon$.

    We wish to show that $\smce(f,y) \gtrsim \varepsilon^{c+1}$. Indeed, let us take
    $\tilde{w}(X) := w(\pi_I(x))$ where $I := [\gamma, 1-\gamma]$, $\pi_I : [0,1] \to I$ is a projection onto the interval $I$, and $\gamma := \varepsilon/C$ for some large constant $C$.

    Note that $\tilde{w}$ is $\Oh(\varepsilon^{-c})$-Lipschitz with respect to the standard metric on $[0,1]$. If $\E (f-y) \tilde{w}(f) \geq \varepsilon/2$, we immediately have $\smce(f,y) \gtrsim \varepsilon^{c+1}$ (we can use $\tilde{w}/L$ as a test function, where $L = \Oh(\varepsilon^{-c})$ is a Lipcshitz constant for function $\tilde{w}$). Otherwise $\E(f-y) (w(f) - \tilde{w}(f)) \geq \varepsilon/2$. Let us call $w_2 := (w - \tilde{w})/2$, such that $\E (f-y)w_2(f) \geq \varepsilon/4$, and moreover $\supp(w_2) \subset [0, 1] \setminus I$, where $w_2$ is $1$-Lipschitz with respect to $d$.

    Since $[0,1] \setminus I$ has two connected components $[0, \gamma)$ and $(1-\gamma, 1]$, on one of those two connected components correlation between the residual $(y-f)$ and $w_2$ has to be at least $\varepsilon/8$. Since the other case is analogous, let us assume for concreteness, that 
    \begin{equation*}
        \E (y - f) w_3(f) \geq \varepsilon/8,
    \end{equation*} where $w_3(x) = w_2(x)$ for $x \in [0, \gamma)$ and $w_3(x) = 0$ otherwise.

    We will show that this implies $\Pr(f \leq \gamma \land y=1) \gtrsim \varepsilon$, and refer to \Cref{clm:smce-lb} to finish the argument.

    Indeed
    \begin{equation*}
        \E (y-f) w_3(f) \leq \E \left[(1-f) \mathbf{1}[f \leq \gamma \land y=1]\right] + \E \left[f \mathbf{1}[f \leq \gamma \land y=0]\right] \leq \Pr(f \leq \gamma \land y=1) + \gamma,
    \end{equation*}
    hence
    \begin{equation*}
        \Pr(f \leq \gamma \land y=1) \geq \varepsilon/8 - \gamma \geq \varepsilon/16,
    \end{equation*}
    where we finally specify $\gamma := \varepsilon/32$.

    To finish the proof, it is enough to show the following
\begin{claim}
\label{clm:smce-lb}
For a random pair $(f,y)$ of prediction and outcome, if $\Pr(f \leq \gamma \land y = 1) \geq \varepsilon$ or $\Pr(f \geq 1 - \gamma \land y = 0) \geq \varepsilon$, where $\gamma = \varepsilon/8$, then $\smce(f,y) \gtrsim \varepsilon^2$.
\end{claim}
\begin{proof}
We will only consider the case $\Pr(f \leq \gamma \land y=1) \geq \varepsilon$. The other case is identical.

Let us take $w(x) := \max(1 - x/2\gamma, 0)$. We have
\begin{equation*}
    \E (y-f) w(f) \geq \frac{1}{2} \Pr(f \leq \gamma \land y=1) - 2 \gamma \Pr(f \leq \gamma \land y=0) \geq \varepsilon/2 - 2 \gamma \geq \varepsilon/4.
\end{equation*}

Since $w$ is $\Oh(1/\varepsilon)$-Lipschitz, we have $\smce(f,y) \gtrsim \varepsilon^2$.
\end{proof}
\end{proof}

\subsection{Sample complexity --- proof of~\Cref{thm:sample-complexity}}

\label{app:sample-complexity}
\begin{lemma}
\label{lem:concentration}
Let $X : [0, 1] \to \R$ be a random function, satisfying with probability $1$, $\|X\|_1 := \int_0^1 |X(t)| \dt \leq 1$ and $\sup_t X(t) \leq \sigma$. Assume moreover that for every $t$, we have $\E[X(t)] = 0$.

Consider now $m$ independent realizations $X_1, X_2, \ldots X_m : [0,1] \to \R$, each identically distributed as $X(t)$, and finally let \begin{equation*}
    \bar{X}(t) := \frac{1}{m} \sum X_i(t).
\end{equation*} 
Then
\begin{equation*}
    \E\left[\|\bar{X}(t)\|_1^2\right] \leq \frac{1}{\sigma m}.
\end{equation*}
\end{lemma}
\begin{proof}
    By Cauchy-Schwartz inequality $\|X\|_1 \leq \|X\|_2 \| \mathbf{1}\|_2 = \|X\|_2$, hence
    \begin{align*}
        \E[\|\bar{X}\|_1^2] & \leq \E[\|\bar{X}\|_2^2] = \E\left[\int \bar{X}(t)^2 \dt \right] \\
        & = \int \E[\bar{X}(t)^2] \dt \\
        & = \frac{1}{m} \int \E[X(t)^2] \dt \\
        & = \frac{1}{m} \E[\|X\|_2^2] \leq \frac{1}{m} \E[\|X\|_1 \|X\|_\infty] \leq \frac{1}{\sigma m} .
    \end{align*}
\end{proof}

\begin{proof}[Proof of~\Cref{thm:sample-complexity}]
Let us first focus on the case $\sigma = \sigma_0$. For a pair $(f, y) \in [0,1] \times \{0, 1\}$,
let us define $X^{(\sigma_0)}_{f,y} : [0,1] \to \mathbb{R}$ as
\begin{equation*}
    X^{(\sigma_0)}_{f,y}(t) := \tilde{K}_{\sigma_0}(f, t)(f-y).
\end{equation*}
Note that $\smECE_{\sigma_0}(\hat{\cD}) = \|\sum_{i} X^{(\sigma_0)}_{f_i, y_i}/m\|_1$, and similarly $\smECE(\cD) = \| \E_{f,y \sim \cD} X^{(\sigma_0)}_{f,y}\|_1$.

Define $\tilde{X}^{(\sigma_0)}_{i} := X^{(\sigma_0)}_{f_i, y_i} - \E_{f,y \sim \cD} X^{(\sigma_0)}_{f,y}$ --- this is a random function, since $(f_i, y_i)$ is chosen at random from distribution $\cD$, and note that:
\begin{enumerate}
    \item Random functions $\tilde{X}^{(\sigma_0)}_i$ for $i \in \{1, \ldots, m\}$ are independent and identically distributed.
    \item With probability $1$, we have $\|\tilde{X}^{(\sigma_00)}_i\|_1 \leq 2\max_f \| \tilde{K}_{\sigma_0}(f, \cdot)\|_1 = 2$.
    \item Similarly, with probability $1$ we have $\|\tilde{X}^{(\sigma_0)}_i\|_\infty \leq 2 \sup_{t_1, t_2} \tilde{K}_{\sigma_0}(t_1, t_2) \leq 2 \sigma_0^{-1}.$
    \item For any $t \in [0,1]$ and $i \in \{1, \ldots, m\}$, we have $\E[\tilde{X}^{(\sigma_0)}_i(t)] = 0$.
\end{enumerate}
Therefore, we can apply~\Cref{lem:concentration} to deduce
\begin{equation*}
    \E \left[\left\|\frac{1}{m} \sum \tilde{X}_i\right\|_1^2\right] \leq \frac{1}{\sigma_0 m},
\end{equation*}
hence, if $m \gtrsim \varepsilon^{-2}\sigma_0^{-1},$ by Chebyshev inequality with probability at least $2/3$ we can bound $\|\sum_i \tilde{X}^{(\sigma_0)}_i / m\|_1 \leq \varepsilon$, and if this event holds, using triangle inequality \begin{equation*}
\smECE_{\sigma_0}(\cD) - \smECE_{\sigma_0}(\hat{\cD})| \leq \|\sum_i \tilde{X}^{(\sigma_0)}_i\|/m \varepsilon.
\end{equation*}

Finally, for $\sigma > \sigma_0$, note that $X^{(\sigma)}_{i} = X^{(\sigma_0)}_{i} \ast \tilde{K}_{N, \sqrt{\sigma^2 - \sigma_0^2}}$ (\Cref{clm:reflected-gaussian-proper}) and therefore as soon as $\|\sum \tilde{X}^{(\sigma_0)}_{i}\| \leq \varepsilon$, we also have
\begin{align*}
    \| \sum_i \tilde{X}^{(\sigma)}_{i} / m\|_1 & = \|\sum_i \tilde{X}^{(\sigma_0)}_{i} \ast \tilde{K}_{N,\sqrt{\sigma^2 - \sigma_0^2}}\|_1 \\
    & \leq \|\sum_i \tilde{X}^{(\sigma_0)}_{i}\|_1 \|\tilde{K}\|_{1\to 1} \leq \varepsilon,
\end{align*}
where
\begin{equation*}
    \|\tilde{K}\|_{1\to 1} := \sup_{t_1} \int_{t_2} |\tilde{K}(t_1, t_2)| \dt \leq 1.
\end{equation*}

This implies $|\smECE_{\sigma}(\cD) - \smECE_{\sigma}(\hat{\cD})| < \varepsilon$ for all $\sigma \geq \sigma_0$.

Finally, if $\smECE_*(\cD) = \sigma_* \geq \sigma_0$, we have $\smECE_{\sigma_*}(\cD) = \sigma_*$, hence $\smECE_{\sigma_*}(\hat{\cD}) \geq \sigma_* - \varepsilon$, and by monotonicity $\smECE_{\sigma_* - \varepsilon}(\hat{\cD}) \geq \sigma_* - \varepsilon$, implying $\smECE_*(\hat{\cD}) \geq \sigma_* - \varepsilon$. Identical argument shows $\smECE_{*}(\hat{\cD}) \leq \sigma_* + \varepsilon.$

\end{proof}

\subsection{Proof of \Cref{thm:consistent-smece-general} \label{sec:proof-of-consistent-smece-general}}
The \Cref{lem:lb} and \Cref{lem:ub} have their correspondent versions in the more general setting where a metric is induced on the space of predictions $[0,1]$ by a monotone function $h : [0,1] \to \R$ --- the proofs are almost identical to those supplied in the special case, except we need to use the more general version of the duality theorem between $\smce$ and $\ldce$, with respect to a metric $d$ (\Cref{thm:duality}).

\begin{lemma}
\label{lem:lb-general}
Let $h$ be an increasing function $h:[0,1] \to \R \cup \{\pm\infty\}$ and $d_h(u,v) = |h(u) - h(v)|$ be the induced metric on $[0,1]$. Let $U\subset \R$ be (possible infinite) interval containing $h([0,1])$ and $K : U \times U \to \R$ be a non-negative symmetric kernel satisfying for every $t_0 \in [0,1]$, $\int K(t_0, t) \dt = 1$, and $\int |t - t_0| K(t, t_0) \dt \leq \gamma$.  Then
\begin{equation*}
    \smce_d(\cD) \leq \smECE_{K, d_h}(\cD) + \gamma.
\end{equation*}
\end{lemma}
The proof is identical to the proof of~\Cref{lem:lb}.

\begin{lemma}
\label{lem:ub-general}
Let $h$ be an increasing function $h : [0,1] \to \mathbb{R} \cup \{\pm \infty\}$, and $d_h(u,v) := |h(u) - h(v)|$ be the induced metric on $[0,1]$.

Let $K$ be a symmetric, non-negative kernel, such that for some $\lambda \leq 1$ and any $t_0, t_1 \in [0,1]$ we have $\int |K(t_0, t) - K(t_1, t)|\dt \leq |t_0 - t_1|/\lambda$. Let $\cD_1, \cD_2$ be a pair of distributions over $[0,1]\times \{0,1\}$. Then
\begin{equation*}
    |\smECE_{h,K}(\cD_1) - \smECE_{h,K}(\cD_2)| \leq \left(\frac{1}{\lambda} + 1\right) W_1(\cD_1, \cD_2),
\end{equation*}
where the Wasserstein distance is induced by the metric $d((f_1, y_1), (f_2, y_2)) = ||h(f_1) - h(f_2)| + |y_1 - y_2|$ on $[0,1] \times \{0,1\}$. 
\end{lemma}
The proof is identical to the proof of~\Cref{lem:ub}.

With those lemmas in hand, as well as the duality theorem for general metric (\Cref{thm:duality}), we can readily deduce~\Cref{thm:consistent-smece-general}. Indeed, \Cref{lem:ub-general} implies that $\smECE_{h, K}(\cD) \leq (1/\lambda + 1)\ldce_{d_h}(\cD)$, and in particular $\smECE_{h, \sigma} \leq (1/\sigma + 1) \ldce_{d_h}(\cD)$.

This means that if $\sigma > 2 \sqrt{\ldce_{d_h}(\cD)}$, then $\smECE_{h, \sigma} \leq \sqrt{\ldce_{d_h}(\cD)}/2 + \ldce_{d_h}(\cD) < \sigma$, and in particular the fixpoint $\sigma_*$ such that $\smECE_{h, \sigma_*}(\cD) = \sigma_*$ needs to satisfy $\sigma_* \leq 2 \sqrt{\ldce_{d_h}(\cD)}$. 

On the other hand, again at this fixpoint $\sigma_*$, using~\Cref{thm:duality} and~\Cref{lem:lb-general}, we have \begin{equation}
    \ldce(\cD) \approx \smce_{d}(\cD) \leq \smECE_{h,\sigma_*}(\cD) + \sigma_* = 2\sigma_*.
\end{equation}
\subsection{Proof of~\Cref{thm:udce}}
Let us consider a distribution $\cD$ over $[0,1] \times \{0,1\}$ and a monotone function $h$, such that $\smECE_{h, \ast} = \sigma_*$.

First, let us define the randomized function $\kappa_1$: let $\pi_{0} : \R \to \R$ be a projection of $\R$ to $h([0,1])$, and let $\eta \sim \mathcal{N}(0, \sigma_*).$ We define \begin{equation*}
    \kappa_1(f) := h^{-1}( \pi_0(h(f) + \eta)).
\end{equation*}
We claim that this $\kappa_1$ satisfy the following two properties:
\begin{enumerate}
    \item $\E_{(f,y) \sim \cD} |d(f, \kappa'(f))| \lesssim \sigma_*$,
    \item $\ECE(\kappa'(f), y) \lesssim \sigma_*$.
\end{enumerate}
Indeed, the first inequality is immediate:
\begin{equation*}
    \E [ d(f, \kappa'(f) ] = \E [ |h(f) - \pi_0(h(f) + \eta)|] \leq \E |\eta| \leq \sigma_*.
\end{equation*}
The proof that $\ECE(\kappa'(f), y) \lesssim \sigma_*$ is identical to the proof of~\Cref{lem:smece_poor}, where such a statement was shown for the standard metric (corresponding to $h(x) = x$).

Finally, those two properties together imply the statement of the theorem: indeed, if $\ECE(f', y) \leq \sigma_*$, we can take $\kappa_2(t) := \E[y | f' = t]$. In this case pair $(\kappa_2(f'), y)$ is perfectly calibrated, and by definition of $\ECE$, we have $\E |\kappa_2(f') - f'| = \ECE(f', y)$. Composing now $\kappa = \kappa_2 \circ \kappa_1$, we have
\begin{equation*}
    \E_{(f,y) \sim \cD} |\kappa(f) - f| \leq \E [ |\kappa_2\circ\kappa_1(f) - \kappa_1(f) |] + \E [|\kappa_1(f) - f|] \lesssim \sigma_*.
\end{equation*}
Moreover distribution $\cD'$ of $(\kappa(f), y)$ is perfectly calibrated.

\subsection{General duality theorem (Proof of~\Cref{thm:duality}) \label{app:duality}}

Let $\mathcal{P} \subset \Delta([0,1] \times \{0,1\})$ be the family of perfectly calibrated distributions. This set is cut from the full probability simplex $\Delta([0,1]\times \{0,1\})$ by a family of linear constraints, specifically $\mu \in \mathcal{P}$ if and only if
    \begin{equation*}
        \forall t, (1-t) \mu(t, 1) - t \mu(t, 0) = 0.
    \end{equation*}
\begin{definition}
Let $\mathcal{F}(H, \R)$ be a family of all functions from $H$ to $\R$.  For a convex set of probability distributions $\mathcal{Q} \subset \Delta(H)$,
we define $\mathcal{Q}^* \subset \mathcal{F}(H, \R)$ to be a set of all functions $q$, s.t. for all $\cD \in \mathcal{Q}$ we have $\E_{x \sim \cD} q(x) \leq 0.$
\end{definition}
\begin{claim}
    The set $\mathcal{P}^* \subset \mathcal{F}([0,1]\times\{0,1\}, \R)$ is given by the following inequalities. A function $H \in \mathcal{P}^*$ if and only if
    \begin{equation*}
    \forall t, \E_{y \sim \mathrm{Ber}(t)} H(t,y) \leq 0.
    \end{equation*}
\end{claim}
\qed

\begin{lemma}
Let $W_1(\cD_1, \cD_2)$ be the Wasserstein distance between two distributions $\cD_1, \cD_2 \in \Delta([0,1] \times, \{0,1\})$ with arbitrary metric $d$ on the set $[0,1]\times \{0,1\}$, and let $Q \subset \Delta([0,1] \times \{0, 1\})$ be a convex set of probability distributions.

The value of the minimization problem
    \begin{align*}
        \min_{\cD_1 \in \mathcal{Q}} W_1(\cD_1, \cD)
    \end{align*}
    is equal to
    \begin{align*}
        \max & \E_{(f,y) \sim \cD} H(f,y) \\
        \text{s.t.} \quad& H \text{ is Lipschitz with respect to $d$,} \\
        & H \in \mathcal{Q}^*.
    \end{align*}
\end{lemma}
\begin{proof}
\newcommand{\AEm}{\mbox{\AE}}

Let us consider a linear space $\AEm$~of all finite signed Radon measures on $X := [0, 1] \times \{0, 1\}$, satisfying $\mu(X) = 0$. We equip this space with the norm $\|\mu\|_{\AEm} := \mathrm{EMD}(\mu_+, \mu_-)$ for measures s.t. $\mu_+(X) = 1$ (and extended by $\|\lambda \mu\|_{\AEm} = \lambda \|\mu \|_{\AEm}$ to entire space). The dual of this space is $\mathrm{Lip}_0(X)$ --- space of all Lipschitz functions on $X$ which are $0$ on some fixed base point $x_0 \in X$ (the choice of base point is inconsequential). The norm on $\mathrm{Lip}_0(X)$ is  $\|W\|_L$ given by the Lipschitz constant of $W$ (see Chapter 3 in \cite{weaver}  for proofs and more extended discussion).

For a function $H$ on $X$ and a measure $\mu$ on $X$, we will write $H(\mu)$ to denote $\int W \,\mathrm{d} \mu$.

The weak duality is clear: for any Lipschitz function $H \in \mathcal{Q}^*$, and any distribution $\cD_1 \in \mathcal{Q}$ we have $H(\cD) \leq H(\cD_1) + W_1(\cD_1, \cD) = W_1(\cD_1, \cD).$

For the strong duality, we shall now apply the following simple corollary of Hahn-Banach theorem.
\begin{claim}[\cite{deutsch}, Theorem 2.5]
Let $(X, \|\cdot\|_X)$ be a normed linear space, $x_0 \in X$, and $P \subset X$ a convex set, and let $d(x, P) := \inf_{p \in P} \|x - p\|_{X}$. Then there is $w \in X^*$, such that $\|w\|_{X^*} = 1$ and $\inf_{p \in P} w(p) - w(x) = d(x, P)$.
\end{claim}
Take a convex set $P \subset \AEm$ given by $P := \{ \cD - q : q \in \mathcal{Q} \}$. Clearly $\min_{D_1 \in \mathcal{Q}} W_1(\cD, \cD_1) = d_{\AEm}(0, P)$ by definition of the space $\AEm$, and hence using the claim above, we deduce 
\begin{equation*}
    d(0, P) = \max_{\tilde{H} \in \mathrm{Lip}_0 : \|\tilde{H}\|_L = 1} \inf_{p \in P} \tilde{H}(p).
\end{equation*}
Taking $\tilde{H}$ which realizes this maximum, we can now consider a shift $H := \tilde{H} - \sup_{q \in \mathcal{Q}} 
\tilde{H}(Q)$, so that $H \in \mathcal{Q}^*$, and verify
\begin{equation*}
    \min_{D_1 \in \mathcal{Q}} W_1(\cD, \cD_1) = d(0, P) = \inf_{p \in P} \tilde{H}(p) = \tilde{H}(\cD) - \sup_{q \in \mathcal{Q}}\tilde{H}(q) = H(\cD).
\end{equation*}

\end{proof}
\begin{corollary}
\label{cor:duality}
    For any metric $d$ on $[0,1] \times \{0, 1\}$, the $\ldce_d(\cD)$ is equal to the value of the following maximization program
    
    \begin{align*}
        \max & \E_{(f,y) \sim \cD} H(f,y) \\
        \text{s.t.} \quad& H \text{ is Lipschitz with respect to $d$} \\
        &\forall t, \E_{y \sim \mathrm{Ber}(t)} H(t,y) \leq 0.
    \end{align*}
\end{corollary}

\begin{lemma}
    For any metric $d$ on $[0,1]$ if we define $\hat{d}$ to be a metric on $[0,1] \times \{0,1\}$ given by $\hat{d}((f_1, y_1), (f_2,y_2)) := d(f_1, f_2) + |y_1 - y_2|$, we have
    \begin{equation*}
        \smce_{d}(\cD) \geq  \ldce_{\hat{d}}(\cD)/2
    \end{equation*}
\end{lemma}
\begin{proof}
We shall compare the value of $\smce_{d}(\cD)$ with the optimal value of the dual as in \Cref{cor:duality}.

Let us assume that for a distribution $\cD$ we have a function $H : [0,1] \times \{0,1\} \to \R$, s.t. $\E_{(f,y) \sim \cD} H(f,y) = \mathrm{OPT}$, which is Lipschitz with respect to $\hat{d}$. We wish to find a function $w: [0,1] \to [-1, 1]$ which is Lipschitz with respect to $d$, s.t.
\begin{equation*}
    \E_{f,y}(f-y) w(f) \geq \mathrm{OPT}/2.
\end{equation*}
Let us take
\begin{equation*}
    w(f) := H(f, 0) - H(f, 1).
\end{equation*}
We will show instead that $w$ is $2$-Lipschitz, $[-1,1]$ bounded and satisfies $\E_{f,y} (f-y) w(f) \geq \mathrm{OPT}$, and the statement of the lemma will follow by scaling.

Let us define $w(f) := H(f,0) - H(f, 1)$. The condition 
\begin{equation*}
    \forall f, \E_{y \sim \mathrm{Ber}(f)} H(f,y) \leq 0
\end{equation*}
is equivalent to $f w(f) \geq H(f, 0)$. Hence
\begin{equation*}
    H(f,y) = y H(f,1) + (1-y) H(f, 0) = H(f, 0) - y w(f) \leq (f-y) w(f),
\end{equation*}
which implies $\E (f-y) w(f) \geq \E H(f,y)$.

Moreover, the function $w(f)$ is bounded by construction of the metric $\hat{d}$ and the assumption that $H(f,y)$ was Lipschitz. Indeed $|w(f)| = |H(f, 0) - H(f, 1)| \leq \hat{d}((f,0), (f, 1)) \leq 1$.
\end{proof}

To finish the proof of~\Cref{thm:duality}, we we are left with the weak duality statement if the distance $d$ on $[0,1]$ satisfies $d(u,v) \geq |u-v|$, we have $\smce_d(\cD) \leq 2\ldce_d(\cD)$. This, as usual, is relatively easy. Let $w$ be a Lipschitz function as in the definition of $\smce_d$, and $\cD'$ be a perfectly calibrated distribution, supplied together with a coupling $\Pi$ between $\cD$ and $\cD'$ such that
\begin{equation*}
    \E_{(f_1, y_1), (f_2, y_2) \sim \Pi} d(f_1, f_2) + |y_1 - y_2| = \ldce_d(\cD),
\end{equation*}
as in the definition of $\ldce_d(\cD)$ (where $(f_1, y_1)$ is distributed according to $\cD$ and $(f_2, y_2)$ according to $\cD'$).

Then
\begin{equation*}
    \left|\E[(f_1 - y_1) w(f_1)]\right| \leq \left|\E[(f_2 - y_2) w(f_2)]\right| + \E[|(f_1 - y_1)(w(f_1) - w(f_2)|] + \E[|(f_1 - f_2 + y_1 - y_2)w(f_1)|].
\end{equation*}
and we can bound those three terms separately: $\E[(f_2 - y_2) w(f_2)] = 0$ since $\cD'$ is perfectly calibrated, 
\begin{equation*}
    \E[|(f_1 - y_1)(w(f_1) - w(f_2))|] \leq \E[d(f_1, f_2)],
\end{equation*}
since $w$ is Lipschitz with respect to $d$ and $|f_1 - y_1| \leq 1$, and
\begin{equation*}
    \E[|(f_1 - f_2 + y_1 - y_2) w(f_1)|] \leq \E[|f_1 - f_2| + |y_1 - y_2|] \leq \E[d(f_1, f_2) + |y_1 - y_2|].
\end{equation*}
Collecting those together we get \begin{equation*}
    \E[(f_1 - y_1) w(f_1)] \leq 2 \E[d(f_1, f_2)] + \E[|y_1 - y_2|] \leq 2 \ldce_{d}(\cD).
\end{equation*}